\pdfoutput=1
\documentclass[11pt]{article}

\usepackage{coling}
\usepackage{algorithmic}

\usepackage[disable]{todonotes} % for uncomments
\usepackage{url}
\usepackage{xspace}
\usepackage{amssymb} % For \checkmark
\usepackage{pifont}  % For \xmark
\usepackage{breakurl}
\usepackage{times}
\usepackage{latexsym}

\usepackage[T1]{fontenc}
\usepackage[utf8]{inputenc}

\usepackage{microtype}

\usepackage{inconsolata}

\usepackage{graphicx}

\usepackage{booktabs}
\usepackage{subcaption}
\usepackage{xspace}
\usepackage{listings}
\usepackage{tcolorbox}
\usepackage{color}
\usepackage{xcolor}
\usepackage{hyperref}
\usepackage{enumitem}
\usepackage{multirow}
\usepackage{multicol}
\usepackage{enumerate}
\usepackage[ruled,vlined]{algorithm2e}  % For writing pseudocode
\usepackage{tikz}
\usetikzlibrary{matrix, positioning}

\title{A Benchmark and Robustness Study of In-Context-Learning with Large Language Models in Music Entity Detection}

\author{Simon Hachmeier \and Robert Jäschke \\
        Berlin School of Library and Information Science \\ Humboldt-Universität zu Berlin \\ \texttt{\{simon.hachmeier,robert.jaeschke\}@hu-berlin.de}}

\begin{document}
%%%%%%%%%%%%%%%%%%%%%%  comments   %%%%%%%%%%%%%%%%%%%%%%

%\definecolor{rv}{rgb}{0.998,0.722,0.635} % Reviewer
\definecolor{sha}{rgb}{0.635,0.998,0.722} % Simon

\definecolor{rja}{rgb}{0.878, 0.831, 0.482} % Robert
\definecolor{msc}{rgb}{0.721, 0.576, 0.862} % Michel
\definecolor{far}{rgb}{0.576, 0.862, 0.592} % Frederik
\definecolor{hsa}{rgb}{0.576, 0.788, 0.862} % Hadi
\definecolor{rv}{rgb}{0.978,0.534,0.534} % Micha

\definecolor{TODO}{rgb}{0.784,0.145,0.00}

\newcommand{\pkomm}[3][TODO]{\todo[color=#1,size=\scriptsize,#2]{\sffamily #1: #3}}
%\renewcommand{\pkomm}[3][TODO]{} % einkommentieren, Kommentare auszublenden
%\newcommand{\komm}[1]{\pkomm{noinline}{#1}}
% hier können Kommentare ausgeblendet werden
% macros for comments: \abbrv inline, \mabbrv in margin

\newcommand{\ffi}[1]{\pkomm[ffi]{inline}{#1}}\newcommand{\mffi}[1]{\pkomm[ffi]{noinline}{#1}} % Frank
\newcommand{\sha}[1]{\pkomm[sha]{inline}{#1}}\newcommand{\msha}[1]{\pkomm[sha]{noinline}{#1}} % Simon
\newcommand{\rja}[1]{\pkomm[rja]{inline}{#1}}\newcommand{\mrja}[1]{\pkomm[rja]{noinline}{#1}} % Robert
\newcommand{\msc}[1]{\pkomm[msc]{inline}{#1}}\newcommand{\mmsc}[1]{\pkomm[msc]{noinline}{#1}} % Michel
\newcommand{\far}[1]{\pkomm[far]{inline}{#1}}\newcommand{\mfar}[1]{\pkomm[far]{noinline}{#1}} % Frederik
\newcommand{\hsa}[1]{\pkomm[hsa]{inline}{#1}}\newcommand{\mhsa}[1]{\pkomm[hsa]{noinline}{#1}} % Hadi
\newcommand{\rv}[1]{\pkomm[rv]{inline}{#1}}\newcommand{\mrv}[1]{\pkomm[rv]{noinline}{#1}} % 

\newcommand{\final}[1]{\textbf{/* for camera ready/long version: #1  */}}

\newcommand{\eg}{e.g.,\xspace}
\newcommand{\ie}{i.e.,\xspace}
\newcommand{\etc}{etc.\xspace}
\newcommand{\csi}{VI \xspace}
\newcommand{\csilong}{Version Identification \xspace}
\newcommand{\shslong}{SecondHandSongs\xspace}
\newcommand{\shs}{SHS\xspace}
\newcommand{\shsK}{SHS100K\xspace}
\newcommand{\mRecoNER}{MusicRecoNER\xspace}
\newcommand{\yt}{YouTube\xspace}
\newcommand{\rd}{Reddit\xspace}
\newcommand{\bert}{BERT\xspace}
\newcommand{\roberta}{RoBERTa\xspace}
\newcommand{\mpnet}{MPNet\xspace}
\newcommand{\mixtral}{Mixtral-8x22B\xspace}
\newcommand{\gpt}{GPT\xspace}
\newcommand{\gptfour}{GPT-4\xspace}
\newcommand{\gptFO}{GPT-4o\xspace}
\newcommand{\gptFOmini}{GPT-4o-mini\xspace}

\newcommand{\phithree}{Phi-3\xspace}

\newcommand{\llama}{Llama\xspace}
\newcommand{\llamatwo}{Llama2\xspace}
\newcommand{\llamathree}{Llama3\xspace}
\newcommand{\llamathreeone}{Llama3.1-70B\xspace}
\newcommand{\llamagroq}{Llama-3-Groq\xspace}
\newcommand{\firefunction}{FireFunction-v2\xspace} 
\newcommand{\nexusraven}{Nexusraven\xspace}
\newcommand{\nuextract}{NuExtract-3B\xspace}

\newcommand{\icl}{ICL\xspace}
\newcommand{\icllong}{in-context-learning\xspace}
\newcommand{\ner}{NER\xspace}
\newcommand{\iex}{IE\xspace}
\newcommand{\ugc}{UGC\xspace}
\newcommand{\llm}{LLM\xspace}
\newcommand{\llms}{LLMs\xspace}
\newcommand{\slm}{SLM\xspace}
\newcommand{\slms}{SLMs\xspace}
\newcommand{\tfidf}{tf-idf\xspace}

\newcommand{\cmark}{\ding{51}}%
\newcommand{\xmark}{\ding{55}}%
\newcommand{\mcorrect}{\cmark}
\newcommand{\mpartial}{(\cmark)}
\newcommand{\mnone}{\xmark}

\newcommand{\iob}{IOB\xspace}
\newcommand{\iobL}{inside-outside-beginning\xspace}
\newcommand{\woa}{WoA\xspace}
\newcommand{\woas}{WoAs\xspace}

\newcommand{\artist}{Artist\xspace}
\newcommand{\artists}{Artists\xspace}

\newcommand{\btag}{B-\xspace}
\newcommand{\itag}{I-\xspace}
\newcommand{\bwoa}{B-\woa}
\newcommand{\iwoa}{I-\woa}
\newcommand{\bartist}{B-\artist}
\newcommand{\iartist}{I-\artist}

\newcommand{\datasetpublic}{MusicUGC-NER\xspace}
\newcommand{\ytdata}{D-YT\xspace}
\newcommand{\rddata}{D-RD\xspace}
\newcommand{\combineddata}{D-RD+YT\xspace}
\newcommand{\fmt}{FMT\xspace}
\newcommand{\synthmemorized}{\fmt Passed\xspace}
\newcommand{\synthunmemorized}{\fmt Failed\xspace}
\newcommand{\synthpostcutoff}{Post-Cutoff\xspace}
\newcommand{\levelone}{Level-1\xspace}
\newcommand{\leveltwo}{Level-2\xspace}

\newcommand{\datasetsizeAll}{78,390\xspace}
\newcommand{\datasetsizeL}{77,858\xspace}
\newcommand{\ntemplates}{1,067\xspace}
\newcommand{\datasetsizeS}{609\xspace}
\newcommand{\datasetsizeJ}{2,977\xspace}
\newcommand{\datasetutts}{1,232\xspace}
\newcommand{\musicutts}{150,000\xspace}
\newcommand{\musicuttsuniq}{38,000\xspace}

% subsets
\newcommand{\perfectmatched}{Both (perfect)\xspace}
\newcommand{\bothmatched}{Both (above threshold)\xspace}
\newcommand{\onlywoa}{Only \woa\xspace}
\newcommand{\onlyartist}{Only Artist\xspace}
\newcommand{\nonematched}{None matched\xspace}

\newcommand{\ytb}[1]{\href{https://youtu.be/#1}{#1}}

%%%%%%%%%%%%%%%%%%%%%%  layout   %%%%%%%%%%%%%%%%%%%%%%
% max. Anzahl an Gleitobjekten auf einer Seite (oben, unten, gesamt)
\setcounter{topnumber}{5}
\setcounter{bottomnumber}{5}
\setcounter{totalnumber}{20}
\renewcommand{\topfraction}{1}
\renewcommand{\bottomfraction}{1}
\renewcommand{\textfraction}{0}
\renewcommand{\floatpagefraction}{0.99}
\def\dbltopfraction{.99}
\def\dblfloatpagefraction{.99}

%%%%%%%%%%%%%%%%%%%%%% squeezing space   %%%%%%%%%%%%%%%%%%%%%%
% warning: use of these tricks is highly discouraged!
% \addtolength{\textfloatsep}{-4mm}
% \addtolength{\dbltextfloatsep}{-.4cm}
% \addtolength{\abovecaptionskip}{-2mm}
% \addtolength{\baselineskip}{-1pt}
% \addtolength{\textwidth}{1mm}
% \addtolength{\textheight}{3mm}

%%% Local Variables:
%%% mode: latex
%%% TeX-master: "paper"
%%% End:

\maketitle
\begin{abstract}

Detecting music entities such as song titles or artist names is a useful application to help use cases like processing music search queries or analyzing music consumption on the web. Recent approaches incorporate smaller language models (SLMs) like BERT and achieve high results. However, further research indicates a high influence of entity exposure during pre-training on the performance of the models. With the advent of large language models (LLMs), these outperform SLMs in a variety of downstream tasks. However, researchers are still divided if this is applicable to tasks like entity detection in texts due to issues like hallucination. In this paper, we provide a novel dataset of user-generated metadata and conduct a benchmark and a robustness study using recent LLMs with in-context-learning (ICL). Our results indicate that LLMs in the ICL setting yield higher performance than SLMs. We further uncover the large impact of entity exposure on the best performing LLM in our study. 

\end{abstract}

\section{Introduction}
\label{sec:intro}

% Fragen
% Contribution 1. & 2.
% - Quantization & Sizes
% - 
% Contribution 3.
% - Generalisierung
% - Unschärfe (Levensthein-basierte Änderung) Generalisierung/Abstraktion
% - Seen/Unseen
% - Datensätze: 1. Mein Datensatz (known), 2. Generiert durch statistisches Modell (unknown) 3. evtl. kleinerer Datensatz (echte Artists und Songs nach dem Knowledge Cut)  
% 

The detection of music entities (\eg song titles and artists) in texts on the web can be of elementary use in various applications such as processing (conversational) search queries \cite{liljeqvist2016nqueries, epure2023human} or the analyses of music consumption on online video platforms. Within these use cases of named entity recognition (\ner) in the music domain, the utterances typically originate from user-generated content (\ugc). 

The difficulties of \ner in \ugc have already been identified, for example, by \citet{jijkoun2008named} and \citet{porcaro2019recognizing}: users can express themselves freely, resulting in potential misspellings or abbreviated utterances of named entities. In the music domain a major challenge arises, which is also common in other creative content domains (\eg movies, books or video games): Unlike other entity classes, such as names of persons, there is no known regular structure or defined vocabulary from which music entities are composed of \cite{derczynski2017results, brasoveanu2020media}. This renders utterances of musical entities susceptible to ambiguity. This phenomenon is not limited to cross-domain ambiguity (\eg the term \emph{Queen} as a band in contrast to the term representing a monarch), but also encompasses class discrimination within the music domain (\eg the album \emph{Queen} by the singer Nicki Minaj). 

The currently proposed state-of-the-art approaches in \ner are mostly based on encoder-only models like \bert \cite{devlin2018bert} and \roberta \cite{liu2019roberta}. 
Although these have been shown to struggle with the aforementioned difficulties, higher exposure of entities in pre-training consequently leads to significantly higher performance of those in testing \cite{lin2020rigorous,epure2022probing,epure2023human}. As a result, some approaches focus on contextual triggers within the context \cite{lin2020triggerner,ma2021enhanced}.

Large language models (\llms), such as \gptfour \cite{openai2024gpt4short} or \llamathree \cite{dubey2024llama3herdmodelsshort},  have been shown to master a variety of natural language tasks. In the case of \ner, researchers are still divided if \llms are the preferred choice for the task in contrast to smaller language models (\slms)\footnote{We adopt the terminology to distinguish between \slms and \llms from \citet{Ma2023large}.} like \roberta, due to problems like hallucination \cite{Ma2023large,sun2023pushing,zhang2024linkner}. However, due to the usually much larger amount of pre-training data of \llms in contrast to \slms, the likelihood of music entity exposure during the pre-training is even higher. This strengthens the question about the performance of \llms in the task as well as the ability to generalize to unseen entities.

In this paper, we aim to address these questions by conducting a benchmark study on a novel dataset of music entities in \ugc using multiple recent \llms with \icllong (\icl). In the second step, we conduct a controlled experiment to investigate the robustness of a selected \llm in which we show discover factors harming its performance. In summary, our contributions are twofold: 
\begin{itemize}
    \item We present an annotated dataset \emph{\datasetpublic} for \ner in the music domain based on user-generated content from the web. We release our dataset publicly\footnote{\url{https://github.com/progsi/YTUnCoverLLM}}\rv{Github Link mit Zenodo Marke + Huggingface Veröffentlichung. Lizenz von MusicRecoNER abschauen.} consisting of Reddit posts provided by \citet{epure2023human}\mrv{die beiden als Co-Authoren auf Zenodo?} and \yt video titles annotated by us. On our proposed dataset we conduct a benchmark comparing fine-tuned \slms with \llms using \icl for the task of \ner of music entities in \ugc. 
    \item From our \ugc dataset, we generate clozes to evaluate the robustness of \llms with regards to unseen music entities and perturbations (\eg typos and abbreviations) in the entity utterances.
\end{itemize}

The remainder of this paper is organized as following: in the next section, we outline related work regarding \ner in the music domain and the task of information extraction, which is a broader task encompassing \ner. In Section~\ref{sec:data} we document our dataset creation procedure and present the corresponding descriptive statistics. In Section~\ref{sec:robustness} we describe methodology to quantify exposure of \llms and our data synthesis to generate new data using our cloze dataset. We describe our experimental design in Section~\ref{sec:experiments} and the respective results in Section~\ref{sec:results}. We close the paper with the conclusion in Section~\ref{sec:conclusion} and reflect limitations of this study in Section~\ref{sec:limitations}.

\section{Related Work}
\label{sec:related}

\paragraph{\ner in the music domain}
Various works proposed \ner approaches to detect music entities like musical artists and song or album titles in (user-generated) texts. Before the broad use of pre-trained language models, \ner approaches were based on conditional random fields (CRF) \cite{liljeqvist2016nqueries, porcaro2019recognizing} or automated voting approaches \cite{oramas2016elmd}. 
\citet{porcaro2019recognizing} propose an approach based on long short-term memory networks (LSTMs) together with CRFs for named entity recognition of classical musical entities. The data is also \ugc since it is gathered from tweets to a radio channel profile. While this dataset also concerns the music domain, it is noteworthy that music entities in classical music are usually more regular than in western pop music, because they follow a structure (\eg \emph{Symphony No.~5} or \emph{Symhony No.~9}).

With the rise of pre-trained \slms, these were widely adapted for \ner generally and in the music domain. \cite{xu2022gazetteer} combine \bert \cite{devlin2018bert} in a mixture-of-experts approach with convolutional neural networks (CNNs) and LSTMs to improve upon the dataset of \cite{porcaro2019recognizing}. \citet{liu2021crossner} proposed pre-training and fine-tuning approaches with \bert to address cross-domain \ner. Their created dataset comprises the music domain and four other domains and is based on Wikipedia articles. Another dataset by \citet{epure2023human} focuses on the use case of conversational music recommendation. The dataset contains user requests for music suggestions on Reddit. We use this dataset and provide a joint dataset together with our annotated data as described in the following section. %The baseline \slms \bert, \roberta and \mpnet achieve F1 scores up to 0.76 which is similar to the one by \citet{liu2021crossner} in the music domain subset. 

%\cite{liljeqvist2016nqueries} proposed a probabilistic approach and a conditional random field with context words.

%\cite{porcaro2019recognizing} propose long short term memory-based approaches together with CRFs for named entity recognition of classical musical entities in radio channel data. They also make use of Twitter data.

%Another dataset is provided by \cite{oramas2016elmd} which is created by an automated voting approach. The code is publicly available.\footnote{\url{https://github.com/sergiooramas/elvis/tree/master}}

\paragraph{\iex with LLMs}

The task of \iex deals with the automatic extraction of relevant structured information from unstructured text. Thus, it is a broader task which encompasses \ner, but also other tasks such as relation extraction. Recently, \llms are applied to a range of different \iex problems. The strategies of \llm use incorporate zero- or in-context-learning (\icl) \cite{wang2023gpt, ashok2023promptner, jung2024llm, ma2024informationHistoric, hachmeier2024ie_music_queries}, auxiliary use in combination with \slms \cite{Ma2023large, peng2024metaie, ye2024llm, zhang2024linkner, zhou2024universalner}, fine-tuning \cite{li2023label}, or reinforcement learning \cite{huang2023adaptive, ding2024adaptive}.

In this paper, we employ a \tfidf-based few-shot prompting based on \citet{hachmeier2024ie_music_queries}, which has shown to be successful for music entities. 
The retrieval of similar few-shot examples to the items in the inference stage was used by other authors as well. \citet{wang2023gpt} achieve near state-of-the-art performance in \ner with \gpt-3 and a few-shot prompting approach where the closest examples for each unseen sample are retrieved with nearest neighbor search. Similarly, \citet{ashok2023promptner} achieve high \ner performance using \gpt-3.5 and \gpt-4. It is noteworthy that some authors state that \llms are still not outperforming \slms in the task due to the increased output space and problems such as hallucination \cite{Ma2023large, sun2023pushing, zhang2024linkner}. 

\citet{sun2023pushing} propose various ideas to mitigate this, such as self-verification and a few-shot demonstration retrieval. Other authors favor the auxiliary use of \llms together with \slms. \citet{Ma2023large} propose to use \llms only for re-ranking the \slm outputs; since they claim that \llms are of better use for hard samples than easy ones. Similarly, \citet{zhang2024linkner} only utilize \llms to re-label uncertain \slm predictions. 
Other techniques include the use of \llms for data augmentation \citet{ye2024llm} or model distillation \cite{zhou2024universalner, peng2024metaie}.

% \paragraph{Quantifying Memorization}

% \cite{carlini2019secret}
% \cite{carlini2021extracting}
% \cite{carlini2023quantifying}
% \cite{schwab2019buster}
% \cite{peshterliev2020self}
% \cite{conroy2023quantifying}

%\citet{ye2024llm} propose LLM-DA, a data augmentation method based on different strategies, to improve the NER task with \bert and \roberta as base models.
%
%Lastly, some works propose model distillation by using LLMs \cite{zhou2024universalner, peng2024metaie}.

%\cite{li2024knowcoder} model universal information extraction as a code generation task. 

%\cite{goel2023llms} applies LLMs for medical information extraction.

\section{Data}\label{sec:data}

Our goal is to benchmark \llms for music entity detection in \ugc. In this section, we describe how we created a novel dataset of \yt video titles containing music entities. 
Our dataset is provided in the \iobL (\iob) format \cite{ramshaw1995textchunking} where each text represents a \yt video title and the tags are referring to entity mentions of either a work of art, such as song titles and albums titles (\woa), or performing artists (\artist). We later join our dataset with \mRecoNER \cite{epure2023human} to cover two different types of \ugc, namely online video metadata and Reddit posts.

\subsection{Dataset Creation}
\label{sec:creation}

\paragraph{Data Sources}
\label{sec:data_sources}

Our dataset contains a subset of works from \shsK \cite{xu2018keyinvariant} which is a large collection of cover songs of (mostly western) popular music. A key advantage for using \shsK is its carefully curated metadata on the platform Secondhandsongs (\shs),\footnote{\url{https://secondhandsongs.com/}} provided by community volunteers. Each cover song is represented by rich information, such as a song title (\woa), an artist name, a composer, a release year, and a link to a \yt video containing a performance of the respective song. This knowledge base serves to test \llms for factual knowledge in Section~\ref{sec:fact_test}. 

To obtain the \ugc utterances, we crawl the corresponding video titles from YouTube under consideration of the fair use policy.\footnote{\href{https://support.google.com/youtube/answer/9783148}{see: Google Support Answer no. 9783148}} From the original \shsK dataset we retain 89,763 representations of videos which are still available on \yt at the time of dataset creation. 76\% of the representations are in the training set from the initial split from \citet{xu2018keyinvariant} and the remaining 24\% account for approximately half of the initial validation and test set. We annotate an approximately equal amount of the initial train, validation and test subsets. Since we can make use of the song-level metadata from \shs, we decided to apply an automatic matching to make the annotation process more efficient. We provide details about the respective pre-processing and matching steps in Appendix~\ref{app:automatic_matching}.

%Before we match the ground truth song attributes from \shs with the utterances in the \ugc, we first conduct several pre-processing steps.

%Hence, each item in the dataset is a song represented by song-level attributes (\artist and \woa) and the \ugc of the \yt video.

%To distinguish between video titles and descriptions, we represent each independently. This results in Z \msha{add number} samples in our dataset where each sample is either a video title or a video description. 

\paragraph{Human Annotation}
\label{sec:human_annotation}

We further obtain annotations by two annotators from our organization and one author. In our annotation tool (see Appendix~\ref{app:annot_tool}) we show the \woa and \artist variations obtained in our pre-processing step in Section~\ref{app:preprocessing} from \shs. The annotators can then select the respective IOB tags per token in a drop-down menu. We provide more details on the annotation protocol in Appendix~\ref{app:annotation_protocol}. 

In total, we obtain \datasetsizeS annotated items, each with two annotators which yield very high agreement (Cohen's Kappa of 0.93 on average). In the following, we refer to our annotated dataset as \ytdata. 

\subsection{Joining with \mRecoNER}
\label{sec:join_mreconer}

Additionally, we join our data with the \mRecoNER dataset \cite{epure2023human}. This dataset is based on Reddit queries by users requesting music recommendations. We use the unprocessed Reddit queries, since their pre-processing includes removal of various special characters, as the authors focused on a conversational use case rather than web data. Since \mRecoNER only contains \iob tags for the processed dataset, we must relabel the word sequences. To achieve this, we align the unprocessed and processed word sequences by matching every word in the unprocessed sequence to the respective word in the processed sequence based not only on the word itself, but also its relative position. The resulting gaps, which are special characters, are labeled based on the two surrounding tags. For instance, if an unlabeled token $x$ in the unprocessed sequence is surrounded by tags of one class and one utterance (\eg \emph{\bwoa $x$ \iwoa} or \emph{\iwoa $x$ \iwoa}), $x$ is assigned an inside tag of the same class (here \iwoa). In all other cases, $x$ is assigned an outside tag.
We refer to this dataset as \rddata and to the combined dataset of the latter with \ytdata as \combineddata. For five-fold cross validation we stratify both datasets to five subsets each representing approximately the same ratio of \yt texts to Reddit texts and we ensure that the same \woas and \artists only occur in one of the subsets together.

\subsection{Dataset Statistics}
\label{sec:statistics}

As can be seen in Table~\ref{tab:dataset}, \woa utterances appear to be longer than \artist utterances. We additionally checked the number of representations without \woa and \artist utterances in the curated subset which account for 6\% and 23\% respectively. 

\begin{figure}
    \centering
    \includegraphics[width=\linewidth]{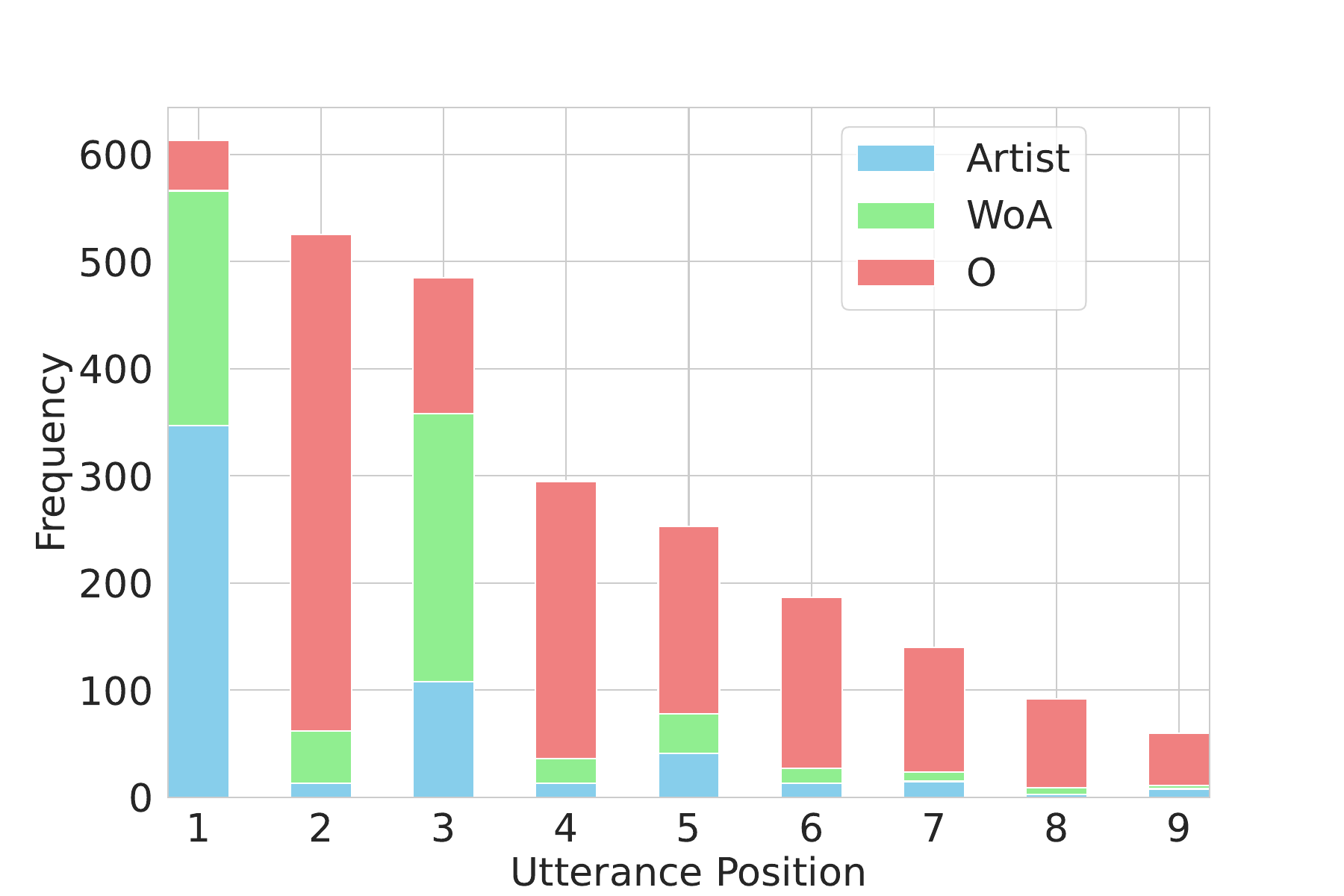}
    \caption{Relative positions of the utterances per class in \ytdata up to the 9th utterance. \emph{O} refers to the outside tag in the \iob format.}
    \label{fig:hist_starts}
\end{figure}

Figure~\ref{fig:hist_starts} shows the relative position of utterances regarding the classes in \ytdata. We observe that video uploaders mostly mention the artist before the \woa. The second utterance is neither \woa nor \artist in the majority of cases, which can indicate the use of a separator (\eg a dash) before the \woa. However, the utterance order is widely spread as we will see later.

\begin{table}
\centering
\begin{tabular}{@{}lrrrrr@{}}
\toprule
& & \multicolumn{2}{c}{\textbf{Words}} & \multicolumn{2}{c}{\textbf{Entities}} \\
Dataset & Items & WoA & Artist & WoA & Artist \\
%\midrule
%Full    & \datasetsizeAll & 3/19 & 2/10 & 1/5 & 1/7 \\
\midrule
\ytdata & \datasetsizeS & 3/15 & 2/9 & 1/3 & 1/4 \\
\combineddata & \datasetsizeJ & 2/15 & 2/9 & 0/5 & 0/7   \\
\bottomrule
\end{tabular}
\caption{Statistics (median/maximum) of the words per entity utterance (\emph{Words}) and the entity utterances per sample (\emph{Entities}) for \ytdata and \combineddata.}
\label{tab:dataset}
\end{table}

\section{Robustness Study}\label{sec:robustness}

%\rv{RV1 Punkt 7: add Definition for \emph{cloze task} or \emph{cloze text}. Evtl. auch \emph{cloze template}?}

We aim to evaluate the robustness of \llms using our dataset. To isolate the surrounding contexts from the entities and to ensure the same class sizes for seen and unseen entities, we additionally perform an experiment on a synthesized dataset using the best performing \llm \gptFOmini.

As discussed in Section~\ref{sec:intro}, the exposure of entities in pre-training can have a significant impact on the performance of \slms. Thus, we focus on the robustness with regard to unseen entities. Since our domain concerns \ugc which is prone to peculiarities such as typos, we further investigate the robustness of the \llm towards perturbations.

\subsection{Quantifying Exposure}

% \paragraph{Wikidata}

% Since both \slms and \llms are trained with data from the web, we argue that more popular music entities are more likely and more often included in the pre-training data. We use the number of sitelinks in Wikipedia, which represent the number of Wikipedia articles in different languages of a given entity. This measure was used as a proxy for popularity before \cite{schwab2019buster, peshterliev2020self, conroy2023quantifying}. 

% Based on the number of sitelinks, we obtain a ranking of all entities in which more sitelinks represent higher popularity and a lower rank. Based on these ranks we are able to compute the exposure metric \cite{carlini2019secret} similar to \citet{epure2023human}. Lasty, based on the exposure metrics we distinguish between seen and unseen entities by a threshold set at 1.5 based on the distribution of exposure. More details on threshold selection, data linking and Wikidata crawling are provided in Appendix~\ref{app:wikidata}. 
%\paragraph{Testing Factual Memorization}
%\label{sec:fact_test}

\paragraph{Factual Memorization Test}
\label{sec:fact_test}

Using the rich \shs metadata as described in Section~\ref{sec:data_sources}, we are able to construct a test to model factual memorization of \llms as defined by \citet{hartmann2023sok}.
In the following, we refer to this test as \fmt.\rv{RV1 Punkt 4. FMT ändern zu \emph{MemTest}? Ausschreiben wäre unpraktisch. Da ich die Abkürzung viel nutze.} 

\begin{figure}[tbph]
    \centering
    \begin{tikzpicture}[node distance=1.5cm, font=\footnotesize]

    % Define styles for boxes
    \tikzstyle{question} = [rectangle, draw, fill=blue!10, text width=10em, align=center, rounded corners, minimum height=2em]
    \tikzstyle{answer} = [rectangle, draw, fill=orange!10, text width=8em, align=center, rounded corners, minimum height=2em]

    % Define styles for headers
    \tikzstyle{header} = [font=\small\bfseries, align=center]

    % Define nodes for questions
    \node (q1) [question] {
        \textbf{Q1} \\
        Who is or who are the original performing artist(s) of the song \textbf{Yesterday} released in the year \textbf{1965}?
    };

    \node (q2) [question, below=0.5cm of q1] {
        \textbf{Q2} \\
        Who wrote the original song of the cover version \textbf{Yesterday} performed by \textbf{Tomcats} in the year \textbf{1966}?
    };

    % Define nodes for answers
    \node (a1_correct) [answer, right=0.5cm of q1] {
        \textit{The Beatles}
    };
    
    \node (a2_correct) [answer, right=0.5cm of q2] {
        \textit{John Lennon, Paul McCartney}
    };

    % Add headers
    %\node (header_q) [header, above=.25cm of q1] {Question};
    \node (header_a) [header, above=.5cm of a1_correct] {Expected Answer};

\end{tikzpicture}
    \caption{Questions of our factual memorization test (\fmt) on the example of the musical work \emph{Yesterday} originally performed by The Beatles.}
    \label{fig:memorization_test}
\end{figure}
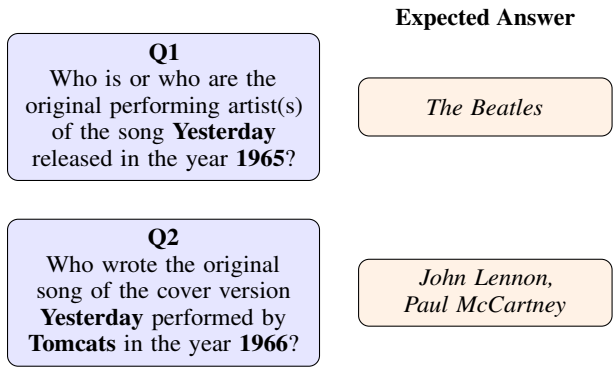

In our domain of cover songs, factual knowledge can be modeled on the level of a musical work. We regard a musical work as a group of cover songs. By definition, a musical work has a composer and an original version with a corresponding original artist \cite{yesiler2021audio}. Based on these two attributes, we can model factual knowledge as tuples with a subject (the work), a relationship and an object, for instance: \emph{(Yesterday; composed\_by; John Lennon, Paul McCartney)}. Based on the relationships to original artists and composers of musical works, we construct our \fmt with two questions as shown in Figure~\ref{fig:memorization_test}. Based on the outcomes of the test with regard to the two questions, we distinguish between three \fmt outcomes:
\begin{description}
    \item[Passed \mcorrect] Both questions answered correctly. 
    \item[Partial \mpartial] One question is correctly answered or at least one answer contains an artist entity which is a performing artist of any cover of the work.
    \item[Failed \mnone] All other outcomes.
\end{description}

We conduct the \fmt for all entities in \ytdata, due to the necessity of entity links to \shs. We provide more details about the implementation of the factual memorization test in Appendix~\ref{app:fact_test}. Next, we also retrieve a subset of real-world music entities that we use for data synthesis. 

\paragraph{Debut Artists}

Beside relying solely on our data, which might be memorized by our used \llms even if the means in the previous sections indicate otherwise, we use music entity information of works which are released after the knowledge cutoff of \gptFOmini which is at the end of 2023. We sample random entities per cloze and denote the resulting synthesized dataset as \synthpostcutoff. We use the API of MusicBrainz\footnote{\url{https://musicbrainz.org/doc/MusicBrainz_API}} to obtain 100 international debut artists of the year 2024 with their debut \woa released. More details about the crawling of MusicBrainz is supplied in Appendix~\ref{app:musicbrainz}. 
%In the following, we describe our data synthesis procedure.

% \begin{table*}
% \begin{tabular}{@{}ll@{}}
% \toprule
% \textbf{Question} & \textbf{Answer} \\ \midrule
% \textbf{Q1} & \\
% Who is (who are) the original performing artist(s) of the song OTITLE released in the year OYEAR? & \\ \midrule
% \textbf{Q2} & \\
% Who performed the song CTITLE on CRELEASE in the year CYEAR? & \\ \midrule
% \textbf{Q3} & \\
% Who wrote the original song of the cover version CTITLE performed by CARTIST in the year CYEAR? & \\ \bottomrule
% \end{tabular}
% \caption{Questions and Answers}
% \label{tab:questions}
% \end{table*}

\subsection{Data Synthesis}
\label{sec:synthesis}

We synthesize data of \ugc in the music domain based on our joint dataset described in Section~\ref{sec:join_mreconer}. To investigate the impact of context tokens which surround the entity utterances, we first create clozes in the \iob format.

\paragraph{Cloze Dataset}
\label{sec:synthesis_templates}

A cloze is a text where some words are masked. In natural language processing, cloze tasks are used to train language models to predict the masked words. In our case, we create clozes from our dataset to construct templates of surrounding context which can be applied to different music entities.

In all texts, we replace the full utterances by a one-word mask per class. Thus, all words which represent \btag or \itag are replaced. For example, the sequence \emph{songs like bohemian rhapsody} is replaced by \emph{songs like [WoA]}. Using this strategy, we can distinguish the isolated surrounding contexts independently of entity mentions. Furthermore, we replace years with a mask, which we detect with regular expressions.\footnote{We match words with four characters and starting with \emph{19} or \emph{20}.} After these steps and the removal of utterances without any entity mention (only outside tags), we obtain \ntemplates unique clozes. We use these clozes to synthesize data by filling in music entities from randomly sampled \woas with corresponding \artists from two of the \fmt outcomes which we denote as \synthmemorized, and \synthunmemorized respectively. Additionally, we create a dataset using the entities from \synthpostcutoff. As a result, we obtain three datasets of the same size using all of the unique clozes. In the next step, we use these three datasets to create perturbed versions.

%\sha{Groups: Post-Knowledge Cutoff, Seen (passed test) and unseen (test not passed). For each, we use all  \ntemplates templates.}

%\sha{Split: Few-Shot Set/Test Set: nur 0Shot vs. stratified}

\paragraph{Perturbations}%\msha{evtl. Bsp. angeben für Operationen mit Musik Entitäten}

%\rv{RV1 Punkt 8c: "The paper does not provide enough detail on how perturbations were added, how words were chosen for perturbation, or how the perturbation probability was calculated."}

As described before, certain perturbations are not unlikely in the case of \ugc. We model different types of perturbations in the utterances of music entities, namely character-level and word-level perturbations of entity mentions similar to \citet{feng2024llmefichecker}. Table~\ref{tab:perturb_example} shows  corresponding examples: Word-level perturbation modifies the sequence of words by either deleting or shuffling tokens. Character-level perturbation alters the characters within a word by performing deletions, insertions, or substitutions. We additionally consider a third type of perturbation which addresses abbreviations which are sometimes used in artist utterances. To model abbreviations, we simply use the first character per word in the artist string. 
\begin{table}[h]
\centering
\begin{tabular}{lll}
\toprule
\textbf{Level} & \textbf{Operation} &  \textbf{Example} \\
\midrule
Input & None & \emph{johnny b goode} \\
Character & Deletion       & \emph{jonny b goode} \\
Character & Insertion       & \emph{johnny b goodey} \\
Character & Substitution       & \emph{johnni b goode} \\
Word & Deletion     & \emph{johnny b} \\
Word & Shuffle     & \emph{johnny goode b} \\
\bottomrule
\end{tabular}
\caption{Example of perturbations per type for the \woa \emph{Johnny B. Goode}.}
\label{tab:perturb_example}
\end{table}
We create two perturbed datasets, which we denote as \levelone and \leveltwo. We set the perturbation probability to $p=0.5$. 
In \levelone with a probability of $p$ we apply one randomly selected perturbation out of two (word-level or character-level). In \leveltwo, we apply up to two perturbations each with a probability of $p$. The first time, it is either an abbreviation perturbation or a word-level perturbation. The second time, it is a 
character-level perturbation. For more details about the perturbation implementation, please refer to our repository. %We now describe our experimental setup.

\section{Experimental Design}
\label{sec:experiments}

With regards to our first contribution, we conduct benchmarks on \ytdata and \combineddata using five-fold cross validation. We ensure that the same entities are only occurring in one fold. In \combineddata, we stratify the folds to gather approximately the same ratio of items from \ytdata and \rddata. For the robustness study described in Section~\ref{sec:robustness}, we use all \ntemplates templates for each of the three synthesized datasets and split the synthesized sets two-fold into test and few-shot sets during experiments to avoid using the exact same clozes in both sets. With the identifier on music entity level, we ensure that utterances of the same music entity do not occur in both sets. In the following, we first describe the  \slm and \llm models used for our benchmarks.

\subsection{\ner with \slms}

We fine-tune \roberta \cite{liu2019roberta} and \bert \cite{devlin2018bert} for the sequence labeling task. Hence, each of the two masked language models transforms an input sequence into a sequence of \iob tag predictions. We use the training parameters as proposed by \citet{epure2023human}. Since we observed that training more epochs was beneficial, we trained the models on 5 epochs instead of 2.

\subsection{\ner-like \iex with \llms}
\label{sec:ie_llm}

\paragraph{Instruction} 

We use \llms with a prompt validated in a previous study \cite{hachmeier2024ie_music_queries}. Rather than mapping the input texts to a sequence of \iob tags like in sequence labeling, we use \llms with \icl in an \iex fashion\footnote{We experimented with sequence labeling, but found that the \llm outputs for sequence labeling were generally not very reliable. It is noteworthy that sophisticated methods can counteract this issue \cite{dukic2024looking}.} and extract information into structured output format. 
The output formats depend on the \llm and are either JavaScript Object Notation (JSON) \cite{pezoa2016foundations} or Pydantic \cite{Pydantic2024}. The key \emph{utterance} maps to the detected music entity in exactly its uttered form which might include potential typos and abbreviations. The key \emph{label} maps to the class of the music entity, namely \woa or \artist. This way, we are able to match the utterances with the input texts to obtain a sequence of labels like in \ner. 
In previous works, several authors discovered the relevance of detailed attribute explanations to effectively leverage \llms in \iex \cite{wang2023gpt, ashok2023promptner, zhang2023promptner}. Therefore, we include detailed attribute explanations. We provide more details about the prompt from \cite{hachmeier2024ie_music_queries} in Appendix~\ref{app:few_shot_prompt}.

\paragraph{Sampling of Few-Shot Examples}

We use the training split as a few-shot example dataset. At each iteration, $k$ few-shot examples are sampled to be included in the prompt. We employ a \tfidf-sampling approach that retrieves most similar examples to the current sample which was shown to be superior to simply random sampling \citet{hachmeier2024ie_music_queries}. %We compare two different methods: random sampling and \tfidf based sampling (cf \ref{app:few_shot_prompt}).

\paragraph{\llms}

We benchmark the following \llms:
\begin{description}
     \item[\firefunction] A model based on \llamathree-70B but optimized for function calling. The authors claim that its performance in benchmarks related to function calling tasks is comparable to \gptFO \cite{garbacki2024firefunction}. We use the Ollama commit \emph{b1ed6b22fb67}.\footnote{\url{https://ollama.com/library/firefunction-v2}}
    \item[\gptFOmini] The smaller version of the most recent flagship model of OpenAI \cite{HelloGPT4o} advancing in performance over the prior series of GPT models \cite{openai2024gpt4short}. We use the version \emph{gpt-4o-mini-2024-07-18}.\footnote{\url{https://platform.openai.com/docs/models/gpt-4o-mini}}
    \item[\llamathreeone] The 70B version of the most recent iteration of \llama models \cite{dubey2024llama3herdmodelsshort}. We use the Ollama commit \emph{c0df3564cfe8}.\footnote{\url{https://ollama.com/library/llama3.1:70b}}
    %\item[\llamagroq \cite{llama3groq}] A version of the 8B  \llamathree optimized for tool use, which is particularly helpful for structured output generation. The pre-training data is the same as for the \llama but it is not fully disclosed.
    \item[\mixtral ] A larger version of the model proposed by \citet{jiang2024mixtralexperts}. It follows the mixture of experts (MoE) paradigm \cite{mistral2024mixtral} and was the best model in our previous study \cite{hachmeier2024ie_music_queries}. We use the Ollama commit \emph{e8479ee1cb51}.\footnote{\url{https://ollama.com/library/mixtral:8x22b}} 
    %\item[\nuextract] A lightweight model of which we use the 3B version. It is a fine-tuned version of \phithree \cite{abdin2024phi3short} specialized on information extraction  \cite{2024nuextract}.

\end{description}

For reproducibility, we set the temperature parameter to 0 for all experiments and use the respective default parameters when using \llms via the OpenAI API and Ollama, respectively. The output format is defined in Pydantic schema for \gptFOmini and \mixtral and in JSON for \llamathreeone and \firefunction. 

\subsection{Metrics}

To measure the overall benchmark results, we consider the F1 scores per entity class (\woa and \artist) and their macro average in the strict evaluation scheme \cite{segura2013semeval9}.

In the robustness study on our dataset, we focus on three erroneous outcomes of \ner as defined by \citet{batista2018named}. Given an example text with \emph{the beatles} as actual \artist and \emph{yesterday} as actual \woa entities, we show the three outcomes:
\begin{description}
    \item[Incorrect] Correct entity boundaries but incorrect type or correct type but incorrect boundaries (\eg \emph{the beatles} as \woa or \emph{yesterday} as \artist).
    \item[Spurious] Neither boundaries nor type matches (\eg \emph{beatles} as \woa).
    \item[Missed] Entity not matched at all (\eg \emph{the beatles} assigned with two outside tags).
\end{description}

%In the next section, we present the results of our benchmark study and generalization experiment.
\section{Results}
\label{sec:results}

\subsection{Benchmark}
\label{sec:results_benchmark}

Table~\ref{tab:benchmark} shows the benchmark results. We observe that the \llm performance increases at least up to $k=15$ few-shot samples for all models. This is especially due to the increased performance in detecting \woa entities which appear to be generally harder to detect than \artist entities. 

The performance of the baseline \slms is higher for \bert than for \roberta, but generally lower than the performance of \llms, especially when comparing to \gptFOmini and \firefunction.

The best model is \gptFOmini, which in the zero-shot setting almost competes with all other models in few-shot settings, but the highest performance is achieved with $k=35$. Thus, we decide to use \gptFOmini in our robustness study of which we present the results in Section~\ref{sec:results_robustness}.

\begin{table}
    \centering
    \begin{tabular}{@{}llcc@{}}
        \toprule
        \multirow{2}{*}{\llm} & \multirow{2}{*}{$k$} & \ytdata
         & \combineddata \\
       \multicolumn{1}{c}{} & \multicolumn{1}{c}{} & \scriptsize Artist/WoA/Avg. & \scriptsize Artist/WoA/Avg. \\
        \midrule
                \multirow{6}{*}{\firefunction} & 0 & .76/.67/.71 & .78/.68/.73 \\
        &  5 & .83/.78/.80 & .82/.70/.80 \\
        &  15 & .86/.81/\textbf{.84} & .84/.78/.81 \\
        &  25 & .84/.81/.82 & .85/.78/.81 \\
        &  35 & .85/.80/.82 & .86/.79/.82 \\
        \midrule
        \multirow{6}{*}{\gptFOmini} & 0 & .85/.78/.81 & .86/.75/.81 \\
        &  5 & .86/\textbf{.82}/\textbf{.84} & \textbf{.88}/.77/.82 \\
        &  15 & \textbf{.87}/.81/\textbf{.84} & \textbf{.88}/.78/.83 \\
        &  25 & .86/.81/\textbf{.84} & \textbf{.88}/.79/.83 \\
        &  35 & .85/\textbf{.82}/\textbf{.84} & \textbf{.88}/\textbf{.80}/\textbf{.84} \\
        \midrule
        \multirow{6}{*}{\llamathreeone} & 0 & .81/.78/.79 & .82/.70/.76 \\
        &  5 & .82/.81/.81 & .82/.74/.78 \\
        &  15 & .84/.81/.83 & .84/.76/.80 \\
        &  25 & .83/\textbf{.82}/.83 & .84/.76/.80 \\
        &  35 & .82/.81/.82 & .84/.75/.80 \\
        \midrule
        \multirow{6}{*}{\mixtral} & 0 & .81/.68/.75 & .73/.67/.80 \\
         &  5 & .83/.79/.81 & .84/.75/.79 \\
         &  15 & .83/.80/.82 & .86/.78/.82 \\
         &  25 & .83/.80/.82 & .86/.78/.82 \\
         &  35 & .83/.80/.81 & .86/.78/.82 \\
        \midrule
        \roberta & - & .78/.72/.75 & .78/.74/.76 \\
        \bert & - &  .82/.74/.79  & .80/.73/.76 \\
        \bottomrule
    \end{tabular}
    \caption{Mean F1 scores (a/b/c) for a) \artist, b) \woa, and c) macro average (Avg.) between a) and b) using the strict evaluation scheme \cite{segura2013semeval9} on the datasets using five-fold cross-validation. Highest results are marked in bold.}
    \label{tab:benchmark}
\end{table}

In Table~\ref{tab:recall_combined} we report the results per subset based on the results of the \fmt introduced in Section~\ref{sec:fact_test}. We observe that the recall of \woa recognition drops by a large margin and up to .24 for \gptFOmini when comparing the performance at works with passed \fmt and failed \fmt. The effect is less apparent when comparing the results of the outcome \emph{partial}.
However, we can already observe a drop in recall here for the models \mixtral and \gptFOmini. The results indicate that in fact the factual knowledge has an impact on the \ner performance, similarly like in the case of \slms. Hence, we further investigate the \llm more closely by our synthesized data.

%This is especially the case for \llamathreeone and \firefunction.\msha{add \gptFOmini} This is not only the case for the subset of works which resulted as \emph{falsely} memorized, but also only \emph{partially} memorized entities. However, when looking at the recall of \artist entities, the degree of memorization does not appear to have an impact. It is noteworthy, that we tested for factual memorization about work entities and not \artist entities directly, which might have an impact on the results. In the next section, we report the results on our synthesized dataset explained in Section~\ref{sec:synthesis}.

\begin{table}
    \centering
    \begin{tabular}{@{}l@{\hskip 6pt}c@{\hskip 6pt}c@{\hskip 6pt}c@{\hskip 6pt}c@{\hskip 6pt}c@{\hskip 6pt}c@{}}
        \toprule
        %\multirow{2}{*}{} & \multicolumn{6}{c}{\textbf{FMT}} \\
         %& \multicolumn{6}{c}{\woa} \\
        %\cmidrule(l){2-7}
         & \multicolumn{2}{c}{\mcorrect} & \multicolumn{2}{c}{\mpartial} & \multicolumn{2}{c}{\mnone} \\
         %& R & S & R & \scriptsize Support & \scriptsize Recall & \scriptsize Support \\
        \midrule
        \mixtral        & .80  & (338)  & .68 & (317) & \textbf{.67} & (46) \\
        \llamathreeone  & .78  & (368)  & .73 & (332) & .61 & (53) \\
        \firefunction   & .69  & (302)  & .68 & (389) & .54 & (60) \\
        \gptFOmini      & \textbf{.85}  & (229) & \textbf{.75} & (420) & .61 & (103) \\
        \bottomrule
    \end{tabular}
    \caption{Recall in detecting \woas and the respective support per outcome of the of the \fmt (Correct: \mcorrect, Partial: \mpartial, and False: \mnone). All reported results are with few-shot settings with $k=35$.}
    \label{tab:recall_combined}
\end{table}

% \begin{table}
%     \centering
%     \begin{tabular}{@{}l@{\hskip 6pt}c@{\hskip 6pt}c@{\hskip 6pt}c@{\hskip 6pt}c@{\hskip 6pt}c@{\hskip 6pt}c@{}}
%         \toprule
%         \multirow{2}{*}{} & \multicolumn{6}{c}{\textbf{Factual Memorization}} \\
%          & \multicolumn{3}{c}{\artist} & \multicolumn{3}{c}{\woa} \\
%         \cmidrule(r){2-4} \cmidrule(l){5-7}
%          & \mcorrect & \mpartial & \mnone & \mcorrect & \mpartial & \mnone \\
%         \midrule
%         \mixtral &  .78 & .84  & .71  & .80  & .68  &  .67 \\
%         \llamathreeone & .75  & .77  & .71  & .78  & .73  & .61 \\
%         \firefunction & .78  & .83  & .77  & .69  & .68  & .54 \\
%         \gptFOmini & .85 & .84  & .84  & .85  &  .75 & .61  \\
%         \bottomrule
%     \end{tabular}
%     \caption{Recall per entity class and degree of factual memorization (Correct: \mcorrect, Partial: \mpartial, and False: \mnone). All reported results are with few-shot settings with $k=35$. \sha{Leave out Artist values? Support Correct/Partial/None: FireFunction (302/389/60), GPT (229/420/103), Llama3.1 (367/332/53), Mixtral (338/317/46)}}
%     \label{tab:recall_combined}
% \end{table}

\subsection{Robustness Study}
\label{sec:results_robustness}

\begin{figure}
    \centering
    \includegraphics[width=\linewidth]{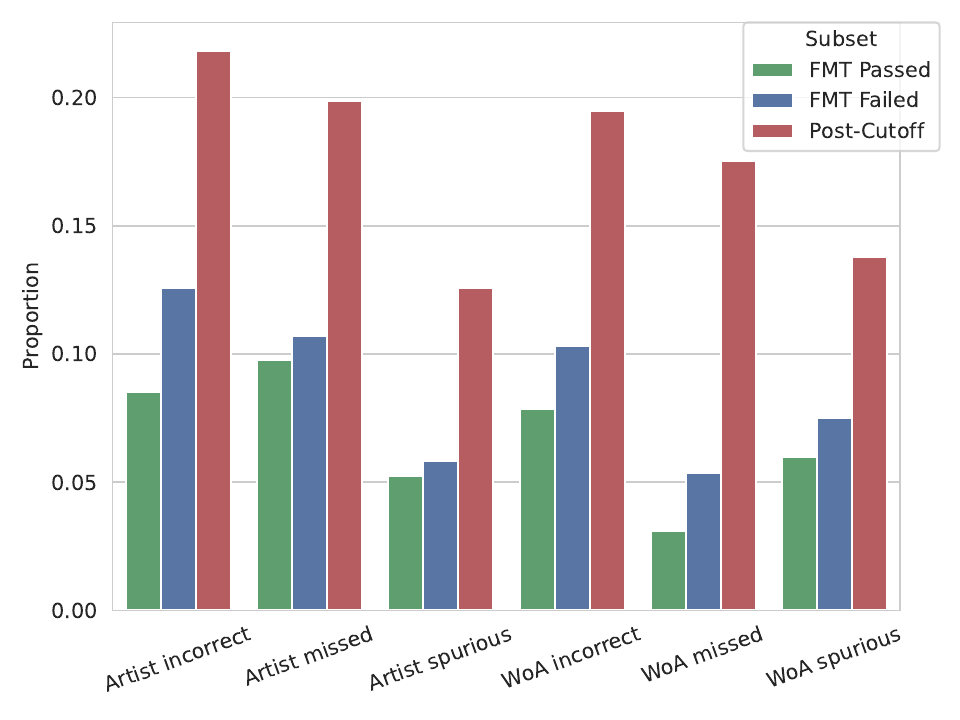}
    \caption{Proportions of errors per group based on our synthesized datasets without perturbation. The total amount is \ntemplates, the number of all unique clozes. }
    \label{fig:error_analysis}
\end{figure}

\begin{figure}
    \centering
    \includegraphics[width=\linewidth]{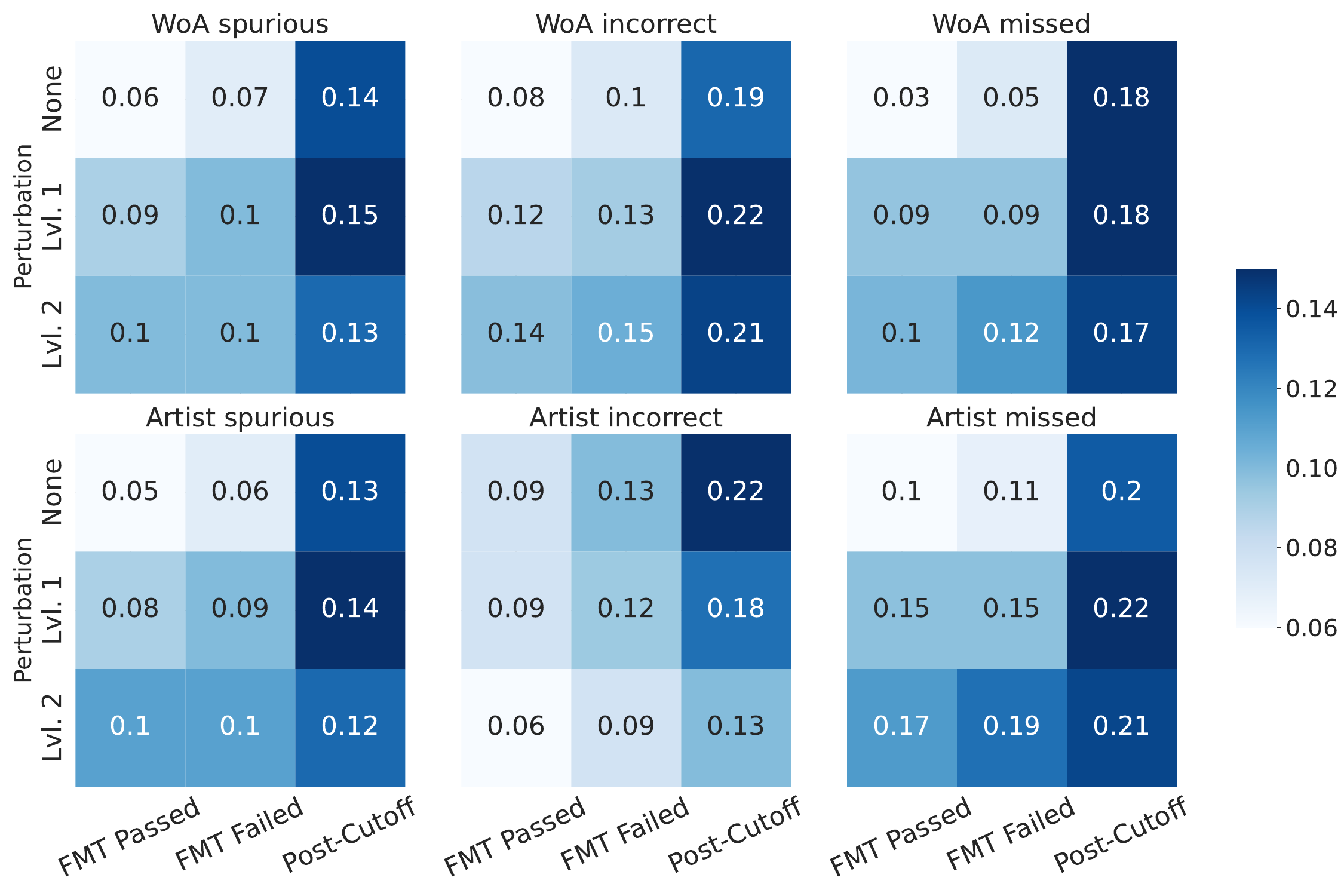}
    \caption{Error proportions per metric per imposed perturbation level and synthesized dataset. The total amount is \ntemplates, the number of all unique clozes.}
    \label{fig:heatmap_groups_perturb}
\end{figure}

We investigate the robustness of the best-performing model in our benchmark, namely \gptFOmini. Figure~\ref{fig:error_analysis} shows the metrics for error analysis. Apparently, the proportions of all errors increase slightly when comparing entities from the passed \fmt with the failed \fmt. However, a much larger difference is visible in case of the entities from \synthpostcutoff. We further see that the most prominent problems are incorrect \artists or \woas as well as missed \artists. In Figure~\ref{fig:heatmap_groups_perturb} we analyze the impact of perturbation and exposure on the ability of robustness of \gptFOmini. In case of the two groups which originate from our dataset and the \fmt groups, increasing the level of perturbation also increases the proportion of errors in almost all cases with the highest increase being for missed artists. Interestingly, in case of \synthpostcutoff the effect is given for all metrics and the perturbation levels decrease the errors for some metrics, such as incorrect \artists. However, this can be due to the dependency of different errors on each other: for example, an actual \woa is predicted as an \artist and hence results in a \emph{missed} \woa and an \emph{incorrect} \artist. Overall, we observe that the effect of exposure appears to be stronger than the effect of perturbation.

\begin{figure}
    \centering
    \includegraphics[width=\linewidth]{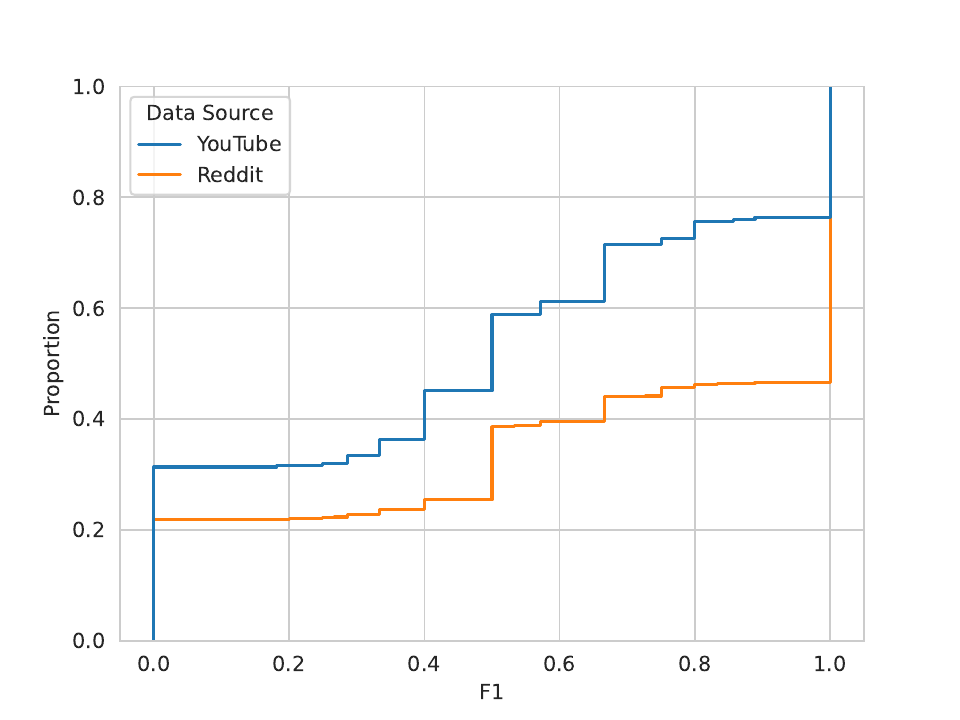}
    \caption{Cumulative distribution functions of F1 scores per data source on our synthesized dataset. }
    \label{fig:cdf_datasource}
\end{figure}

Lastly, we examine the relevance of the surrounding contexts of music entities. In Figure~\ref{fig:cdf_datasource} we compare by the data sources \yt and Reddit for \synthpostcutoff. The distributions indicate that the contexts of Reddit are generally more helpful for the \llm to detect unseen entities. This is probably due to the richer contextual cues in questions (\eg \emph{songs like \dots}) than in online video metadata. However, for both data sources we identified erroneous surrounding contexts (see Table~\ref{tab:outcomes_examples} in Appendix~\ref{app:cloze_texts}). 

\section{Conclusion}
\label{sec:conclusion}

In this paper, we propose a novel dataset for \ner of music entities in user-generated content. The dataset was created using human annotations supported by automatic annotation and additionally joint with the \mRecoNER dataset. In a benchmark experiment, we compared four different \llms in an \icl setting to strong baselines. In our second experiment, we synthesize data to gather more insights about the impact of perturbation and entity exposure on the \llm performance. Our results indicate that \llms with \icl are a strong choice for music entity extraction. However, part of this appears to be due to entity exposure during pre-training. In future studies, logical next steps include the consideration of music gazetteers, which has been pursued in the past \cite{xu2022gazetteer}. In context of \llms, these could be combined with retrieval-augmented generation methods.

\section{Limitations}
\label{sec:limitations}
With regard to our dataset it is noteworthy that the focus lies on western popular music and covers from other regions are not represented as broadly. Furthermore, the genders of artists are hard to determine based on the metadata on Secondhandsongs. Thus, we cannot guarantee gender diversity. 
It is noteworthy that we employed three annotators of our organization which are all working in an academic context on a daily basis. 

In our experiments, we used open-source \llms and one closed-source \llm, namely \gptFOmini. It is possible that the benchmark performance can still be surpassed by other state-of-the-art models, for example, \gptFO or the 405B version of \llamathreeone, which we cannot run due to local resource limitations. For all \llms we relied on the default parameters for quantization and did not test different configurations.

Lastly, we clarify that we purely investigated the performance on our task in focus and did not specifically address the scalability. For instance, both BERT and RoBERTa have less parameters than the tested LLMs. In some cases, the former might still be an attractive alternative, even though they achieve inferior performance on the task.

%With regards to the test of factual memorization, we rely on string matching of pre-processed strings to determine correctness of the answers. It has to be noted that the \llm might still give a correct answer with a minor typo which might result in a wrong answer.

\section*{Acknowledgments}

We thank our student assistants Chris Herrmann and Jonathan Lüpfert for supporting the annotation process.

% Bibliography entries for the entire Anthology, followed by custom entries
%\bibliography{anthology,custom}
% Custom bibliography entries only

%\sha{Quellen von Llama3 und Mixtral: besser mit "Team" abkürzen anstatt der vielen Namen? Sind nicht alle Namen gelistet.}

\bibliography{literature}

\begin{thebibliography}{48}
\providecommand{\natexlab}[1]{#1}

\bibitem[{Ashok and Lipton(2023)}]{ashok2023promptner}
Dhananjay Ashok and Zachary~C Lipton. 2023.
\newblock Promptner: Prompting for named entity recognition.
\newblock \emph{arXiv preprint arXiv:2305.15444}.

\bibitem[{Batista(2018)}]{batista2018named}
David~S. Batista. 2018.
\newblock \href {https://www.davidsbatista.net/blog/2018/05/09/Named_Entity_Evaluation/} {Named-entity evaluation metrics based on entity-level}.
\newblock Accessed: 2024-09-13.

\bibitem[{Brasoveanu et~al.(2020)Brasoveanu, Weichselbraun, and Nixon}]{brasoveanu2020media}
Adrian~MP Brasoveanu, Albert Weichselbraun, and Lyndon Nixon. 2020.
\newblock In media res: a corpus for evaluating named entity linking with creative works.
\newblock In \emph{Proceedings of the 24th Conference on Computational Natural Language Learning}, pages 355--364.

\bibitem[{Colvin et~al.(2023)Colvin, Jolibois, Ramezani, Garcia~Badaracco, Dorsey, Montague, Matveenko, Trylesinski, Runkle, Hewitt, and Hall}]{Pydantic2024}
Samuel Colvin, Eric Jolibois, Hasan Ramezani, Adrian Garcia~Badaracco, Terrence Dorsey, David Montague, Serge Matveenko, Marcelo Trylesinski, Sydney Runkle, David Hewitt, and Alex Hall. 2023.
\newblock \href {https://docs.pydantic.dev/latest/} {Pydantic}.
\newblock If you use this software, please cite it as below.

\bibitem[{Derczynski et~al.(2017)Derczynski, Nichols, van Erp, and Limsopatham}]{derczynski2017results}
Leon Derczynski, Eric Nichols, Marieke van Erp, and Nut Limsopatham. 2017.
\newblock \href {https://doi.org/10.18653/v1/W17-4418} {Results of the {WNUT}2017 shared task on novel and emerging entity recognition}.
\newblock In \emph{Proceedings of the 3rd Workshop on Noisy User-generated Text}, pages 140--147, Copenhagen, Denmark. Association for Computational Linguistics.

\bibitem[{Devlin et~al.(2018)Devlin, Chang, Lee, and Toutanova}]{devlin2018bert}
Jacob Devlin, Ming{-}Wei Chang, Kenton Lee, and Kristina Toutanova. 2018.
\newblock \href {https://arxiv.org/abs/1810.04805} {{BERT:} pre-training of deep bidirectional transformers for language understanding}.
\newblock \emph{CoRR}, abs/1810.04805.

\bibitem[{Ding et~al.(2024)Ding, Ke, Huang, Jiang, Li, Yang, Xiao, and Liang}]{ding2024adaptive}
Zepeng Ding, Ruiyang Ke, Wenhao Huang, Guochao Jiang, Yanda Li, Deqing Yang, Yanghua Xiao, and Jiaqing Liang. 2024.
\newblock Adaptive reinforcement learning planning: Harnessing large language models for complex information extraction.
\newblock \emph{arXiv preprint arXiv:2406.11455}.

\bibitem[{Dubey et~al.(2024)Dubey, Jauhri, Pandey, Kadian, Al-Dahle, Letman, Mathur, Schelten, Yang, Fan, Goyal, Hartshorn, Yang, Mitra, Sravankumar, Korenev, Hinsvark, Rao, Zhang, Rodriguez, Gregerson, Spataru, Roziere, Biron, Tang, Chern, Caucheteux, Nayak, Bi, and et~al.}]{dubey2024llama3herdmodelsshort}
Abhimanyu Dubey, Abhinav Jauhri, Abhinav Pandey, Abhishek Kadian, Ahmad Al-Dahle, Aiesha Letman, Akhil Mathur, Alan Schelten, Amy Yang, Angela Fan, Anirudh Goyal, Anthony Hartshorn, Aobo Yang, Archi Mitra, Archie Sravankumar, Artem Korenev, Arthur Hinsvark, Arun Rao, Aston Zhang, Aurelien Rodriguez, Austen Gregerson, Ava Spataru, Baptiste Roziere, Bethany Biron, Binh Tang, Bobbie Chern, Charlotte Caucheteux, Chaya Nayak, Chloe Bi, and Chris~Marra et~al. 2024.
\newblock \href {https://arxiv.org/abs/2407.21783} {The llama 3 herd of models}.
\newblock \emph{Preprint}, arXiv:2407.21783.

\bibitem[{Duki{\'c} and {\v{S}}najder(2024)}]{dukic2024looking}
David Duki{\'c} and Jan {\v{S}}najder. 2024.
\newblock Looking right is sometimes right: Investigating the capabilities of decoder-only llms for sequence labeling.
\newblock In \emph{Findings of the Association for Computational Linguistics ACL 2024}, pages 14168--14181.

\bibitem[{Epure and Hennequin(2023)}]{epure2023human}
Elena Epure and Romain Hennequin. 2023.
\newblock \href {https://doi.org/10.18653/v1/2023.eacl-main.92} {A human subject study of named entity recognition in conversational music recommendation queries}.
\newblock In \emph{Proceedings of the 17th Conference of the European Chapter of the Association for Computational Linguistics}, pages 1281--1296, Dubrovnik, Croatia. Association for Computational Linguistics.

\bibitem[{Epure and Hennequin(2022)}]{epure2022probing}
Elena~V. Epure and Romain Hennequin. 2022.
\newblock \href {https://aclanthology.org/2022.lrec-1.151} {Probing pre-trained auto-regressive language models for named entity typing and recognition}.
\newblock In \emph{Proceedings of the Thirteenth Language Resources and Evaluation Conference}, pages 1408--1417, Marseille, France. European Language Resources Association.

\bibitem[{Feng et~al.(2024)Feng, Han, Chen, and Yang}]{feng2024llmefichecker}
Xiaoning Feng, Xiaohong Han, Simin Chen, and Wei Yang. 2024.
\newblock \href {https://doi.org/10.1145/3664812} {Llmeffichecker: Understanding and testing efficiency degradation of large language models}.
\newblock \emph{ACM Trans. Softw. Eng. Methodol.}, 33(7).

\bibitem[{Garbacki and Chen(2024)}]{garbacki2024firefunction}
Pawel Garbacki and Benny Chen. 2024.
\newblock \href {https://fireworks.ai/blog/firefunction-v2-launch-post} {Firefunction-v2: {{Function}} calling capability on par with {{GPT4o}} at 2.5x the speed and 10\% of the cost}.
\newblock Accessed: 2024-09-06.

\bibitem[{Hachmeier and Jäschke(2024)}]{hachmeier2024ie_music_queries}
Simon Hachmeier and Robert Jäschke. 2024.
\newblock Information extraction of music entities in conversational music queries.
\newblock In \emph{Proceedings of the 3rd Workshop on NLP for Music and Audio (NLP4MusA)}.

\bibitem[{Hartmann et~al.(2023)Hartmann, Suri, Bindschaedler, Evans, Tople, and West}]{hartmann2023sok}
Valentin Hartmann, Anshuman Suri, Vincent Bindschaedler, David Evans, Shruti Tople, and Robert West. 2023.
\newblock Sok: Memorization in general-purpose large language models.
\newblock \emph{arXiv preprint arXiv:2310.18362}.

\bibitem[{Huang et~al.(2023)Huang, Liang, Li, Xiao, and Ji}]{huang2023adaptive}
Wenhao Huang, Jiaqing Liang, Zhixu Li, Yanghua Xiao, and Chuanjun Ji. 2023.
\newblock \href {https://doi.org/10.18653/v1/2023.findings-acl.863} {Adaptive ordered information extraction with deep reinforcement learning}.
\newblock In \emph{Findings of the Association for Computational Linguistics: ACL 2023}, pages 13664--13678, Toronto, Canada. Association for Computational Linguistics.

\bibitem[{Jiang et~al.(2024)Jiang, Sablayrolles, Roux, Mensch, Savary, Bamford, Chaplot, de~las Casas, Hanna, Bressand, Lengyel, Bour, Lample, Lavaud, Saulnier, Lachaux, Stock, Subramanian, Yang, Antoniak, Scao, Gervet, Lavril, Wang, Lacroix, and Sayed}]{jiang2024mixtralexperts}
Albert~Q. Jiang, Alexandre Sablayrolles, Antoine Roux, Arthur Mensch, Blanche Savary, Chris Bamford, Devendra~Singh Chaplot, Diego de~las Casas, Emma~Bou Hanna, Florian Bressand, Gianna Lengyel, Guillaume Bour, Guillaume Lample, Lélio~Renard Lavaud, Lucile Saulnier, Marie-Anne Lachaux, Pierre Stock, Sandeep Subramanian, Sophia Yang, Szymon Antoniak, Teven~Le Scao, Théophile Gervet, Thibaut Lavril, Thomas Wang, Timothée Lacroix, and William~El Sayed. 2024.
\newblock \href {https://arxiv.org/abs/2401.04088} {Mixtral of experts}.
\newblock \emph{Preprint}, arXiv:2401.04088.

\bibitem[{Jijkoun et~al.(2008)Jijkoun, Khalid, Marx, and De~Rijke}]{jijkoun2008named}
Valentin Jijkoun, Mahboob~Alam Khalid, Maarten Marx, and Maarten De~Rijke. 2008.
\newblock Named entity normalization in user generated content.
\newblock In \emph{Proceedings of the second workshop on Analytics for noisy unstructured text data}, pages 23--30.

\bibitem[{Jung et~al.(2024)Jung, Kim, and Jang}]{jung2024llm}
Sung~Jae Jung, Hajung Kim, and Kyoung~Sang Jang. 2024.
\newblock Llm based biological named entity recognition from scientific literature.
\newblock In \emph{2024 IEEE International Conference on Big Data and Smart Computing (BigComp)}, pages 433--435. IEEE.

\bibitem[{Li et~al.(2023)Li, Li, Liu, Xie, Li, Wang, Li, and Zhong}]{li2023label}
Zongxi Li, Xianming Li, Yuzhang Liu, Haoran Xie, Jing Li, Fu-lee Wang, Qing Li, and Xiaoqin Zhong. 2023.
\newblock Label supervised llama finetuning.
\newblock \emph{arXiv preprint arXiv:2310.01208}.

\bibitem[{Liljeqvist(2016)}]{liljeqvist2016nqueries}
Sandra Liljeqvist. 2016.
\newblock Named entity recognition for search queries in the music domain.

\bibitem[{Lin et~al.(2020{\natexlab{a}})Lin, Lee, Shen, Moreno, Huang, Shiralkar, and Ren}]{lin2020triggerner}
Bill~Yuchen Lin, Dong-Ho Lee, Ming Shen, Ryan Moreno, Xiao Huang, Prashant Shiralkar, and Xiang Ren. 2020{\natexlab{a}}.
\newblock Triggerner: Learning with entity triggers as explanations for named entity recognition.
\newblock \emph{arXiv preprint arXiv:2004.07493}.

\bibitem[{Lin et~al.(2020{\natexlab{b}})Lin, Lu, Tang, Han, Sun, Wei, and Yuan}]{lin2020rigorous}
Hongyu Lin, Yaojie Lu, Jialong Tang, Xianpei Han, Le~Sun, Zhicheng Wei, and Nicholas~Jing Yuan. 2020{\natexlab{b}}.
\newblock \href {https://doi.org/10.18653/v1/2020.emnlp-main.592} {A rigorous study on named entity recognition: Can fine-tuning pretrained model lead to the promised land?}
\newblock In \emph{Proceedings of the 2020 Conference on Empirical Methods in Natural Language Processing (EMNLP)}, pages 7291--7300, Online. Association for Computational Linguistics.

\bibitem[{Liu et~al.(2019)Liu, Ott, Goyal, Du, Joshi, Chen, Levy, Lewis, Zettlemoyer, and Stoyanov}]{liu2019roberta}
Yinhan Liu, Myle Ott, Naman Goyal, Jingfei Du, Mandar Joshi, Danqi Chen, Omer Levy, Mike Lewis, Luke Zettlemoyer, and Veselin Stoyanov. 2019.
\newblock \href {https://arxiv.org/abs/1907.11692} {Roberta: A robustly optimized bert pretraining approach}.
\newblock \emph{Preprint}, arXiv:1907.11692.

\bibitem[{Liu et~al.(2021)Liu, Xu, Yu, Dai, Ji, Cahyawijaya, Madotto, and Fung}]{liu2021crossner}
Zihan Liu, Yan Xu, Tiezheng Yu, Wenliang Dai, Ziwei Ji, Samuel Cahyawijaya, Andrea Madotto, and Pascale Fung. 2021.
\newblock Crossner: Evaluating cross-domain named entity recognition.
\newblock In \emph{Proceedings of the AAAI Conference on Artificial Intelligence}, volume~35, pages 13452--13460.

\bibitem[{Ma and Liu(2021)}]{ma2021enhanced}
Liwen Ma and Weifeng Liu. 2021.
\newblock An enhanced method for entity trigger named entity recognition based on pos tag embedding.
\newblock In \emph{2021 IEEE 7th International Conference on Cloud Computing and Intelligent Systems (CCIS)}, pages 89--93. IEEE.

\bibitem[{Ma et~al.(2023)Ma, Cao, Hong, and Sun}]{Ma2023large}
Yubo Ma, Yixin Cao, Yong Hong, and Aixin Sun. 2023.
\newblock \href {https://doi.org/10.18653/v1/2023.findings-emnlp.710} {Large language model is not a good few-shot information extractor, but a good reranker for hard samples!}
\newblock In \emph{Findings of the Association for Computational Linguistics: EMNLP 2023}. Association for Computational Linguistics.

\bibitem[{Ma et~al.(2024)Ma, Santo, Lackey, Viswanathan, and O'Malley}]{ma2024informationHistoric}
Zhiwei Ma, Javier~E Santo, Greg Lackey, Hari Viswanathan, and Daniel O'Malley. 2024.
\newblock Information extraction from historical well records using a large language model.
\newblock \emph{arXiv preprint arXiv:2405.05438}.

\bibitem[{{Mistral AI Team}(2024)}]{mistral2024mixtral}
{Mistral AI Team}. 2024.
\newblock \href {https://mistral.ai/news/mixtral-8x22b/} {Cheaper, better, faster, stronger continuing to push the frontier of ai and making it accessible to all.}
\newblock Accessed: 2024-09-06.

\bibitem[{OpenAI(2024)}]{HelloGPT4o}
OpenAI. 2024.
\newblock \href {https://openai.com/index/hello-gpt-4o/} {Hello {{GPT-4o}}}.
\newblock Accessed: 2024-09-11.

\bibitem[{OpenAI et~al.(2024)OpenAI, Achiam, Adler, Agarwal, Ahmad, Akkaya, Aleman, Almeida, Altenschmidt, Altman, Anadkat, Avila, Babuschkin, Balaji, Balcom, Baltescu, Bao, Bavarian, Belgum, Bello, Berdine, Bernadett-Shapiro, Berner, Bogdonoff, Boiko, Boyd, Brakman, Brockman, Brooks, Brundage, and et~al.}]{openai2024gpt4short}
OpenAI, Josh Achiam, Steven Adler, Sandhini Agarwal, Lama Ahmad, Ilge Akkaya, Florencia~Leoni Aleman, Diogo Almeida, Janko Altenschmidt, Sam Altman, Shyamal Anadkat, Red Avila, Igor Babuschkin, Suchir Balaji, Valerie Balcom, Paul Baltescu, Haiming Bao, Mohammad Bavarian, Jeff Belgum, Irwan Bello, Jake Berdine, Gabriel Bernadett-Shapiro, Christopher Berner, Lenny Bogdonoff, Oleg Boiko, Madelaine Boyd, Anna-Luisa Brakman, Greg Brockman, Tim Brooks, Miles Brundage, and Kevin~Button et~al. 2024.
\newblock \href {https://arxiv.org/abs/2303.08774} {Gpt-4 technical report}.
\newblock \emph{Preprint}, arXiv:2303.08774.

\bibitem[{Oramas et~al.(2016)Oramas, Anke, Sordo, Saggion, and Serra}]{oramas2016elmd}
Sergio Oramas, Luis~Espinosa Anke, Mohamed Sordo, Horacio Saggion, and Xavier Serra. 2016.
\newblock \href {https://aclanthology.org/L16-1528} {{ELMD}: An automatically generated entity linking gold standard dataset in the music domain}.
\newblock In \emph{Proceedings of the Tenth International Conference on Language Resources and Evaluation ({LREC}'16)}, pages 3312--3317, Portoro{\v{z}}, Slovenia. European Language Resources Association (ELRA).

\bibitem[{Peng et~al.(2024)Peng, Wang, Yao, Wang, and Shang}]{peng2024metaie}
Letian Peng, Zilong Wang, Feng Yao, Zihan Wang, and Jingbo Shang. 2024.
\newblock Metaie: Distilling a meta model from llm for all kinds of information extraction tasks.
\newblock \emph{arXiv preprint arXiv:2404.00457}.

\bibitem[{Pezoa et~al.(2016)Pezoa, Reutter, Suarez, Ugarte, and Vrgo{\v{c}}}]{pezoa2016foundations}
Felipe Pezoa, Juan~L Reutter, Fernando Suarez, Mart{\'\i}n Ugarte, and Domagoj Vrgo{\v{c}}. 2016.
\newblock Foundations of json schema.
\newblock In \emph{Proceedings of the 25th International Conference on World Wide Web}, pages 263--273. International World Wide Web Conferences Steering Committee.

\bibitem[{Porcaro and Saggion(2019)}]{porcaro2019recognizing}
Lorenzo Porcaro and Horacio Saggion. 2019.
\newblock Recognizing musical entities in user-generated content.
\newblock \emph{Computaci{\'o}n y Sistemas}, 23(3):1079--1088.

\bibitem[{Ramshaw and Marcus(1995)}]{ramshaw1995textchunking}
Lance~A. Ramshaw and Mitchell~P. Marcus. 1995.
\newblock \href {https://arxiv.org/abs/cmp-lg/9505040} {Text chunking using transformation-based learning}.
\newblock \emph{Preprint}, arXiv:cmp-lg/9505040.

\bibitem[{Ranjan and Poddar(2022)}]{ranjan2022multilingual}
Ekagra Ranjan and Naman Poddar. 2022.
\newblock \href {https://arxiv.org/abs/2204.01848} {Multilingual abusiveness identification on code-mixed social media text}.
\newblock \emph{Preprint}, arXiv:2204.01848.

\bibitem[{Segura-Bedmar et~al.(2013)Segura-Bedmar, Mart{\'\i}nez, and Herrero-Zazo}]{segura2013semeval9}
Isabel Segura-Bedmar, Paloma Mart{\'\i}nez, and Mar{\'\i}a Herrero-Zazo. 2013.
\newblock \href {https://aclanthology.org/S13-2056} {{S}em{E}val-2013 task 9 : Extraction of drug-drug interactions from biomedical texts ({DDIE}xtraction 2013)}.
\newblock In \emph{Second Joint Conference on Lexical and Computational Semantics (*{SEM}), Volume 2: Proceedings of the Seventh International Workshop on Semantic Evaluation ({S}em{E}val 2013)}, pages 341--350, Atlanta, Georgia, USA. Association for Computational Linguistics.

\bibitem[{Sparck~Jones(1972)}]{sparck1972tfidf}
Karen Sparck~Jones. 1972.
\newblock A statistical interpretation of term specificity and its application in retrieval.
\newblock \emph{Journal of documentation}, 28(1):11--21.

\bibitem[{Sun et~al.(2023)Sun, Dong, Li, Wan, Wang, Zhang, Li, Cheng, Lyu, Wu, and Wang}]{sun2023pushing}
Xiaofei Sun, Linfeng Dong, Xiaoya Li, Zhen Wan, Shuhe Wang, Tianwei Zhang, Jiwei Li, Fei Cheng, Lingjuan Lyu, Fei Wu, and Guoyin Wang. 2023.
\newblock \href {https://arxiv.org/abs/2306.09719} {Pushing the limits of chatgpt on nlp tasks}.
\newblock \emph{Preprint}, arXiv:2306.09719.

\bibitem[{Wang et~al.(2023)Wang, Sun, Li, Ouyang, Wu, Zhang, Li, and Wang}]{wang2023gpt}
Shuhe Wang, Xiaofei Sun, Xiaoya Li, Rongbin Ouyang, Fei Wu, Tianwei Zhang, Jiwei Li, and Guoyin Wang. 2023.
\newblock \href {https://arxiv.org/abs/2304.10428} {Gpt-ner: Named entity recognition via large language models}.
\newblock \emph{Preprint}, arXiv:2304.10428.

\bibitem[{Xu and Qi(2022)}]{xu2022gazetteer}
Wenjia Xu and Yangyang Qi. 2022.
\newblock Gazetteer enhanced named entity recognition for musical user-generated content.
\newblock In \emph{2022 3rd International Conference on Computer Science and Management Technology (ICCSMT)}, pages 40--43. IEEE.

\bibitem[{Xu et~al.(2018)Xu, Chen, and Yang}]{xu2018keyinvariant}
Xiaoshuo Xu, Xiaoou Chen, and Deshun Yang. 2018.
\newblock \href {https://doi.org/10.1109/ICME.2018.8486531} {Key-invariant convolutional neural network toward efficient cover song identification}.
\newblock In \emph{2018 IEEE International Conference on Multimedia and Expo (ICME)}, pages 1--6.

\bibitem[{Ye et~al.(2024)Ye, Xu, Wang, Zhou, Zhang, Gui, and Huang}]{ye2024llm}
Junjie Ye, Nuo Xu, Yikun Wang, Jie Zhou, Qi~Zhang, Tao Gui, and Xuanjing Huang. 2024.
\newblock Llm-da: Data augmentation via large language models for few-shot named entity recognition.
\newblock \emph{arXiv preprint arXiv:2402.14568}.

\bibitem[{Yesiler et~al.(2021)Yesiler, Doras, Bittner, Tralie, and Serr{\`a}}]{yesiler2021audio}
Furkan Yesiler, Guillaume Doras, Rachel~M Bittner, Christopher~J Tralie, and Joan Serr{\`a}. 2021.
\newblock Audio-based musical version identification: Elements and challenges.
\newblock \emph{IEEE Signal Processing Magazine}, 38(6):115--136.

\bibitem[{Zhang et~al.(2023)Zhang, Yan, Zhou, and Qiu}]{zhang2023promptner}
Mozhi Zhang, Hang Yan, Yaqian Zhou, and Xipeng Qiu. 2023.
\newblock Promptner: A prompting method for few-shot named entity recognition via k nearest neighbor search.
\newblock \emph{arXiv preprint arXiv:2305.12217}.

\bibitem[{Zhang et~al.(2024)Zhang, Zhao, Gao, and Hu}]{zhang2024linkner}
Zhen Zhang, Yuhua Zhao, Hang Gao, and Mengting Hu. 2024.
\newblock \href {https://doi.org/10.1145/3589334.3645414} {Linkner: Linking local named entity recognition models to large language models using uncertainty}.
\newblock In \emph{Proceedings of the ACM on Web Conference 2024}, WWW '24, page 4047–4058, New York, NY, USA. Association for Computing Machinery.

\bibitem[{Zhou et~al.(2024)Zhou, Zhang, Gu, Chen, and Poon}]{zhou2024universalner}
Wenxuan Zhou, Sheng Zhang, Yu~Gu, Muhao Chen, and Hoifung Poon. 2024.
\newblock \href {https://arxiv.org/abs/2308.03279} {Universalner: Targeted distillation from large language models for open named entity recognition}.
\newblock \emph{Preprint}, arXiv:2308.03279.

\end{thebibliography}

\appendix

\section{Appendix}
\label{sec:appendix}

\subsection{Automatic Annotation}
\label{app:automatic_matching}

\paragraph{Pre-Processing}
\label{app:preprocessing}

We conduct various pre-processing steps to ensure a robust automatic matching where we match the \shs content against the \ugc from \yt. First, all texts are transformed to lowercase and apostrophes are removed. We then apply specific pre-processing methods for the two data sources due to their peculiarities. In the \shs song-level metadata we found that \woas are often accompanied by additional information in brackets, such as \emph{(acoustic)} or \emph{(remastered)}.
We discard this additional information to just retain the creative content. In case of the artist strings, we consider artists string variations with and without articles (\ie \emph{the beatles} and \emph{beatles}). We found that featuring artists are represented within single strings in the \shs metadata. Hence, we separate the artists to detect these as individual entities. For both, articles and representations for featuring separators (\eg \emph{feat.}), we use pre-defined lists covering multiple languages which we provide in Appendix~\ref{app:prelists}. 
In case of the \yt video titles, we discovered the use of font-like non-Latin texts. To enable matching these texts as well, we perform Unicode normalization similar to \cite{ranjan2022multilingual}. For example, the Unicode character \emph{Mathematical Fraktur Capital F}(code point U+1D509) is normalized to \emph{Latin Capital Letter F} (code point U+0046). 
% https://www.compart.com/de/unicode/U+0046

\subsubsection{Pre-Defined Lists}
\label{app:prelists}
%\rja{einheitlicher sortieren und gruppieren und möglichst immer gleiche Sprachreihenfolge einhalten}

%\paragraph{Articles}
%\label{app:prelists_articles}

We consider the following languages for our dataset, which are contained in the source dataset SHS100K and which represent frequent languages in western popular music: English, French, German, Italian, Portuguese, and Spanish. For each of these languages, we document the articles (\eg \emph{the} for English or \emph{der}, \emph{die} and \emph{das} for German) and different expressions representing separators for featuring artists. For the latter, we consider the form \emph{and + pronoun} (\eg \emph{and her}, for cases like \emph{Billie Holliday and her Orchestra}) and \emph{with + pronoun} (\eg \emph{and her}, for cases like \emph{Billie Holliday with her Orchestra}). The details can be found in the preprocessing script in our repository.

% \begin{itemize}
%     \item the
%     \item le, les
%     \item der, die, das
%     \item il, lo, l', i, gli, le
%     \item a, o, os, as
%     \item la, el, los, las
% \end{itemize}

% \paragraph{\emph{Featuring}-Separators}
% \label{app:prelists_featseps}

% \begin{itemize}
%     \item \emph{and + genitive:}
%     \begin{itemize}
%         \item and his, and her
%         \item et le, et son, et ses, et les
%         \item und sein, und ihr, und seine, und ihre
%         \item e la sua, e la seu, e seu, e sua
%         \item e sua, e a sua
%         \item y su
%     \end{itemize}
%     \item \emph{with} + possessive pronoun:
%     \begin{itemize}
%         \item with her, with his, with the
%         \item avec son, avec sa, avec ses
%         \item mit ihrem, mit ihren, mit seinem, mit seinen
%         \item con sua
%         \item con la sua, con il suo, con i suoi, con le sue 
%         \item com o seu, com o
%         \item con su
%     \end{itemize}
%     \item \emph{featuring}:
%     \begin{itemize}
%     \item featuring, feat, feat., ft. 
%     \end{itemize}
%     \item \emph{and:}
%     \begin{itemize}
%         \item and, et, und, e, y,
%     \end{itemize}
%     \item Chars: /,-\&

% \end{itemize}

\subsubsection{Matching}
\label{app:matching}

We run an algorithm to match the pre-processed variations of song-level attributes \woa and \artist from \shs with the pre-processed \yt metadata texts. We use the partial alignment ratio,\footnote{\url{https://rapidfuzz.github.io/RapidFuzz/Usage/fuzz.htmlpartial-ratio-alignment}} which is the normalized Indel similarity of the optimal alignment of the shorter string to the longer string. Since it is handling alignment of strings of different lengths it is well suited for our use case, where often the utterances occur at different word indices among additional information (\eg \emph{\dots performing yesterday in \dots}). We set the minimum matching threshold to $\tau = 80$ of 100.

Table~\ref{tab:matching_statistics} shows the resulting matching statistics. From the subset of samples where both entities match, in 87\% of the cases both entities are matched with $s=100$. For human annotation, we sample a minimum of 150 representations randomly stratified to subsets shown in Table~\ref{tab:matching_statistics} and the initial subsets of \shsK.  Using the annotated dataset which we obtain in Section~\ref{sec:human_annotation}, we are able to evaluate the automatic matching algorithm, which yields 0.92 in precision and 0.50 in recall. Since the low recall indicates that half of the entities are missed, we only make use of our human annotated dataset in this paper. However, we also provide the automatically matched data in our repository.

\begin{table}[tbph]
    \centering
    \begin{tabular}{@{}lrr@{}}
        \toprule
        \textbf{Matched Entities} & \textbf{Count} & \textbf{Fraction} \\
        \midrule
        Both & 77,889 & 87\% \\ % perfect 67,963
        % $s=100$ 
        % $s \geq \tau$
        \onlywoa & 7,061 & 8\% \\
        % $s_{A} < \tau$
        \onlyartist  & 3,986 & 4\% \\
        % $s_{T} < \tau$
        None & 827 & 1\% \\
        % $s < \tau$ 
        \bottomrule
    \end{tabular}
    \caption{Numbers of samples with matches for both attributes (top) and with at least one non-matching attribute (bottom) based on the similarity $s$ and threshold $\tau=80$.}
    \label{tab:matching_statistics}
\end{table}

\subsection{Annotation Guidelines}
\label{app:annotation_protocol}

The purpose of this annotation task is to label each token in \yt video titles using the \iobL (\iob) format. The task focuses on identifying two specific classes: Artist and \woa (Work of Art). 

\subsubsection{Representation}

Each title is split into individual tokens (words, punctuation marks, etc.). A token typically corresponds to a word, but punctuation marks (e.g., commas, apostrophes) and special characters are treated as separate tokens.
All text is converted to lowercase before annotation.

\subsubsection{Classes}

The following classes are in the focus of this annotation task:

\begin{description}
    \item[Artist] This refers to the name of a musical performing artist or group. It includes singers, bands, DJs, or any individual/group credited for the creation or performance of the music. Group members which do not perform music as individuals are excluded.
    \item[\woa] This refers to the titles of songs, albums, EPs, or any other artistic work related to music.
\end{description}

\subsubsection{\iob Format}

The \iob format was proposed by \citet{ramshaw1995textchunking} and consists of three tags:

\begin{description}
    \item[\btag (Beginning)] Indicates the first token of a named entity.
    \item[\itag (Inside)] Indicates any token that is inside a named entity but not the first one.
    \item[O (Outside)] Indicates tokens that do not belong to any named entity.
\end{description}
    
If a named entity (here: Artist or \woa) consists of a single token, label it with the \btag prefix (\ie \bartist or \bwoa). If a named entity spans multiple tokens, label the first token with the \btag prefix and the subsequent tokens with the \itag prefix (\ie \bartist \iartist). All other tokens not related to Artist or \woa should be labeled as O.

\subsubsection{Ambiguous Cases}

\paragraph{Entity Utterances with Additional Information}

Any additional information, such as regarding the version of the \woa, should be labeled as O. For instance in \emph{the beatles - yesterday (~karaoke version~)}, only \emph{yesterday} should be labeled as \woa. This results in \emph{\bartist \iartist O \bwoa O O O O}.

\paragraph{Ambiguity Between Artist and \woa}

If a token could be interpreted as either an Artist or a \woa, use the surrounding context to make the correct annotation. If context is insufficient, try to find the correct class for the utterance on the web. 

\paragraph{Incorrect Tokens}

In cases where the title contains additional, incorrect tokens within a \woa, annotate all relevant tokens as part of the \woa to maintain the entity's integrity. For example, \emph{nothing else random matters} should be annotated as \emph{B-WoA I-WoA I-WoA I-WoA} to treat the phrase as a single entity despite the inclusion of the irrelevant word \emph{random}. However, if the number of incorrect tokens is large and the entity is not recognizable, you can consider splitting the utterance or just annotating half of the utterance with the corresponding class. Ultimately, this is a case-by-case decision and should depend on your perception of readability.

\paragraph{Featuring Artists}

In titles that include featured artists, both the main artist and the featured artist should be annotated.
For example, \emph{rihanna \mbox{feat . drake}} should be annotated as \emph{\bartist O O \bartist.}

\paragraph{Punctuation Marks}

Punctuation marks should generally be labeled as O unless they are part of the official name of the Artist or \woa.
\emph{p ! ink} should be annotated as \emph{\bartist \iartist \iartist}.

\paragraph{Nested Entity Utterances}

Sometimes entity utterances can be nested. In these cases, favor the outermost entity. For example, \emph{b - sides the beatles} should be labeled as one \woa entity, because the utterance refers to the tribute album \emph{B-Sides The Beatles} from The Beatles.
The only exception are medleys. These shall be annotated as separate \woas. 

\subsubsection{Tool Usage
}
\label{app:annot_tool}

The annotation process will be conducted using our custom annotation tool which was developed using Streamlit.\footnote{\url{https://github.com/streamlit/streamlit}} An example of the tool's interface is shown in Figure 1, illustrating how tokens are presented with corresponding dropdowns for \iob tag selection and how reference information is displayed.

\begin{figure}
    \centering
    \includegraphics[width=\linewidth]{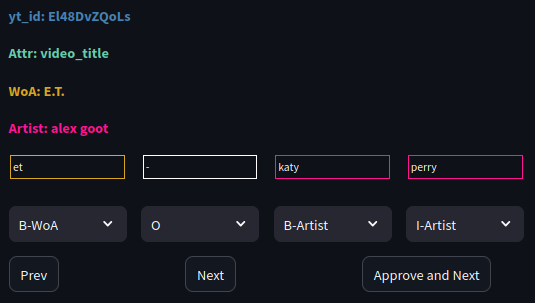}
    \caption{Graphical user interface of our annotation tool. The example shown is a cover of \emph{E.T.} by Katy Perry from the performing artist Alex Goot.}
    \label{fig:annot_gui}
\end{figure}

\paragraph{Dropdown Selection}

In the tool, each token of the video title is displayed with a dropdown menu underneath it. Annotators should use these dropdowns to assign the appropriate IOB tags \bartist, \iartist, \bwoa, \iwoa, O).

\paragraph{Reference Information}

To assist with accurate annotation, the tool also displays one correct Artist and \woa (Work of Art) that are relevant to the YouTube video. This information is supplementary, since also other \woas or \artists shall be annotated.

\subsubsection{Examples}

Some examples for correct annotations are given in Table~\ref{tab:annot_examples}.

\begin{table}[h]
\noindent\rule{\linewidth}{1pt}

% Example 1
\begin{tabular}{cccc}
adele & - & hello \\
\bartist & O & \bwoa \\
\end{tabular}

\vspace{0.4cm} % Add some space between examples

% Example 2
\begin{tabular}{ccccc}
the & beatles & - & hey & jude \\
\bartist & \iartist & O & \bwoa & \iwoa \\
\end{tabular}

\vspace{0.4cm} % Add some space between examples

% Example 3
\begin{tabular}{cccccc}
rihanna & feat & . & drake & - & work \\
\bartist & O & O & \bartist & O & \bwoa \\
\end{tabular}

\vspace{0.4cm} % Add some space between examples

% Example 4
\begin{tabular}{ccccccc}
taylor & swift & - & red & ( & deluxe & )  \\
\bartist & \iartist & O & \bwoa & O & O & O  \\
\end{tabular}

\noindent\rule{\linewidth}{1pt}

\caption{Examples for correct annotations.}
\label{tab:annot_examples}
\end{table}

\subsection{Factual Memorization Test}
\label{app:fact_test}

\begin{figure}[tbph]
    \centering
    \includegraphics[scale=0.5]{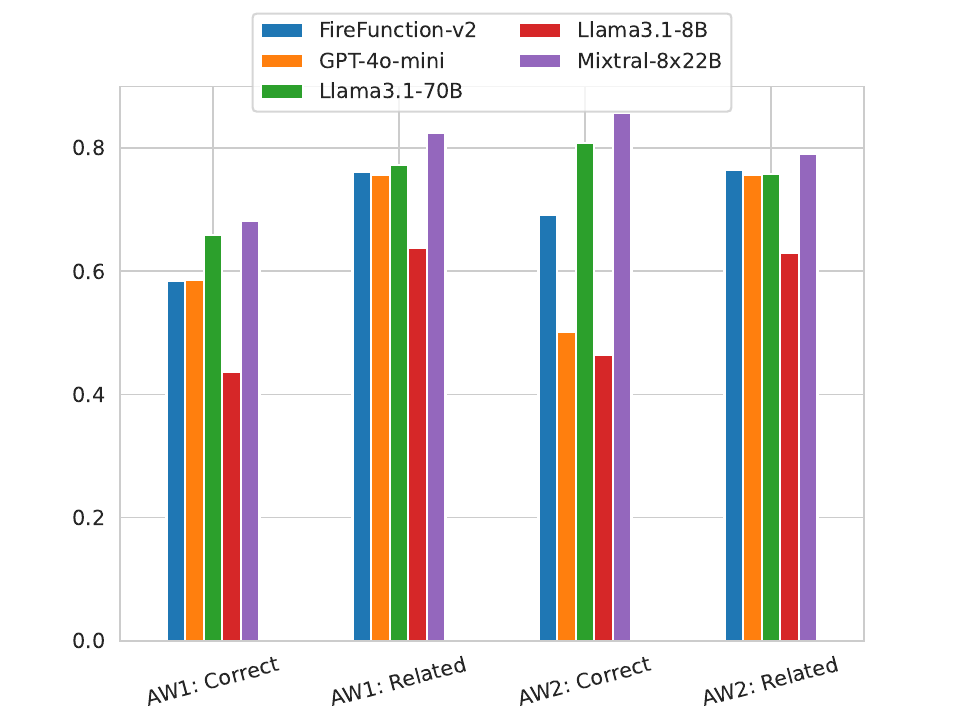}
    \caption{Fractions of correctly answered questions in the memorization test. We distinguish between \emph{correct} and \emph{related} where the artist/composer mentioned is related (\eg a covering artist) to the work, but not strictly correct.}
    \label{fig:memorization_test_results}
\end{figure}

Figure~\ref{fig:memorization_test_results} provides an overview of the outcomes of the factual memorization test using \llamathreeone. Further, we provide some details about our definition of correctness.

%\paragraph{Correctness}
We model the correctness of an answer as a matching string to the ground truth attribute from \shs. Since multiple strings (artists or composer) can be correct answers (\eg Paul McCartney and John Lennon are composers of the song \emph{Yesterday}), we decided that one correct answers is sufficient. For the string matching step, we apply the same pre-processing techniques as described in Section~\ref{app:preprocessing}.

\subsection{Debut Artists from MusicBrainz}
\label{app:musicbrainz}

Using the MusicBrainz API, we crawl for artists with the query text \emph{begin:2024}. We limit the results to the first 100. We show some examples of the artists and corresponding debut releases in Table~\ref{tab:debut_works}.

\begin{table*}
\centering
\begin{tabular}{@{}lll@{}}
\toprule
\textbf{Artist} & \textbf{Debut WoA} & \textbf{Release Date}  \\
\midrule
Ben Keller & Fake Yøu øut & 02-06-2024 \\
The Houseboat Tapes & The Houseboat Tapes & 03-13-2024 \\
Human Fade & Getter & 01-06-2024  \\
Green Buffalo Steak Ensemble & 001: Mary Juana Had a Little Lamb & 05-03-2024  \\
%el hombre que miraba al cielo & se escondió donde el sol & (unknown)  \\
Veskraunghulthyr & Lornmyr Frost & 02-28-2024  \\
I:mond & WE ARE GRAVITY & 06-04-2024 \\
Hitori Kakurenbo & Maquetas & 01-27-2024  \\
\bottomrule
\end{tabular}
\caption{Examples of debut artists in 2024 and their debut \woas retrieved from MusicBrainz.}
\label{tab:debut_works}
\end{table*}

%\subsection{Templates}
\label{app:templates}

\subsection{In-Context-Learning}
\label{app:few_shot_prompt}

\paragraph{Prompt}

Figure~\ref{fig:prompt} shows an example of a prompt to the \llms with sampled few-shot examples. 
We also found that the use of a third label as a wildcard (\emph{other}) is helpful to improve LLM performance, since in many cases with no true entities the models still labeled utterances as \emph{\woa}. Beside utterance and label attributes, we request contextual cues.

\begin{figure}
\small
 \centering
    \begin{tcolorbox}[colback=white!95!gray, colframe=black!75]
    \textbf{Instruction}
    
    From the following text, which contains a user request for music suggestions, extract all the relevant entities that you find.\\
    
    \textbf{Entity Attributes}
    \begin{itemize}[nosep, left=0pt, labelsep=5pt]
        \item \textbf{utterance}: The utterance of the entity in the text. For example ``the beatles'' in ``recommend me music like the beatles''. An utterance can only be of a type for which labels are defined.
        \item \textbf{label}: The label of the entity. It can either be `TITLE' (if the utterance refers to a song or album name), `PERFORMER' (if the utterance refers to a performing artist) or `OTHER' for any other entity type.
        \item \textbf{cue}: The contextual cue which indicates the entity (\eg ``music like'' in ``recommend me music like the beatles'' indicating ``the beatles'')
    \end{itemize}     
    \vspace{10pt}    
    \textbf{Examples} \\
       Input: \textcolor{purple}{stuff like flylo}\\
       (\{'utterance': 'flylo', 'label': 'performer', 'cue': ''\})\\
       Input:  \textcolor{purple}{dré anthony brand new} \\
       \dots
       
    \textbf{Output Schema} 
    
    \dots
    
    \textbf{Input}
    
    \textcolor{darkblue}{songs similar to black bird by alter bridge}
    \vspace{0.5em}
    \end{tcolorbox}

 \caption{Prompt with \textcolor{purple}{few-shot examples} and \textcolor{darkblue}{input text}.}
  \label{fig:prompt}
\end{figure}

\paragraph{\tfidf Sampling}

\begin{figure}
    \centering
    \includegraphics[width=\linewidth]{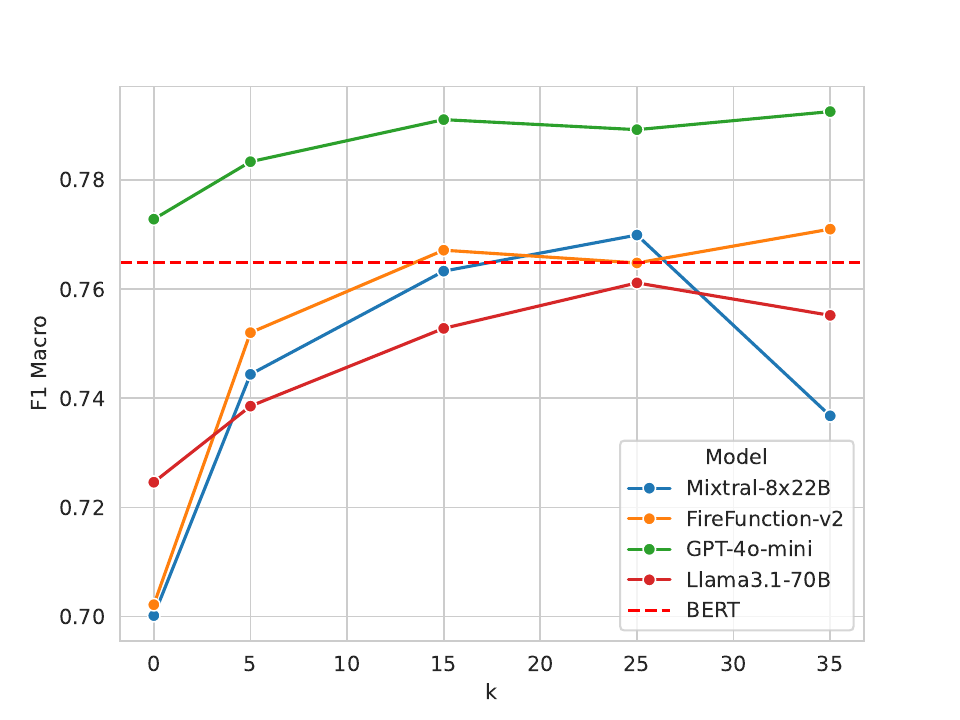}
    \caption{Performance on \combineddata using random sampling as opposed to \tfidf-sampling.}
    \label{fig:results_rand_sampling}
\end{figure}

Term frequency inverse document frequency (\tfidf) \cite{sparck1972tfidf} is a measure of the importance of words in a document in information retrieval. We obtain \tfidf vectors for texts during few-shot sampling. To focus on the similarity of the syntactical structure of the \ugc utterances rather than the content of the tokens of the attributes, we mask the entities when computing the \tfidf similarity. For instance, for the example \emph{songs like nothing else matters by metallica} we get \emph{songs like [WoA] by [Artist]}. The prompt contains the actual texts as shown in Figure~\ref{fig:prompt}. We compared to random sampling \cite{hachmeier2024ie_music_queries} which turned out to yield inferior performance. Figure~\ref{fig:results_rand_sampling} shows the corresponding performance of random sampling for different values of $k$ on \combineddata.

% \begin{figure}
%     \centering
%     \begin{minted}{python}
%     from pydantic import BaseModel
    
%     class MusicEntityV1(BaseModel):
%         """
%         Data model of a music entity
%         """
%         utterance: str
%         label: str
%         cue: str

%     class MusicEntityV2(BaseModel):
%         """
%         Data model of a music entity
%         """
%         utterance: str
%         utterance_normalized: str
%         label: str
%         separator: str
%     \end{minted}
%     \caption{Schema Definitions}
%     \label{fig:schema1}
% \end{figure}

\subsection{Clozes}
\label{app:cloze_texts}

\begin{figure}%[tbph]
    \centering
    \includegraphics[scale=0.5]{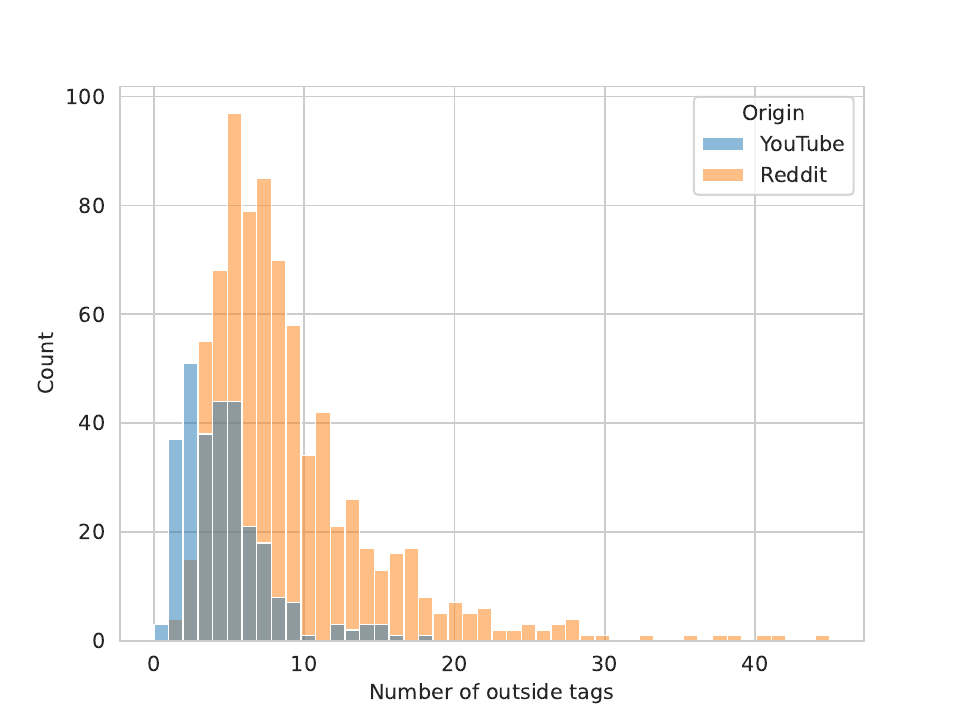}
    \caption{Distribution of the number of outside tokens.}
    \label{fig:hist_outside_tokens_templates}
\end{figure}

\begin{figure}
    \centering
\includegraphics[width=\linewidth]{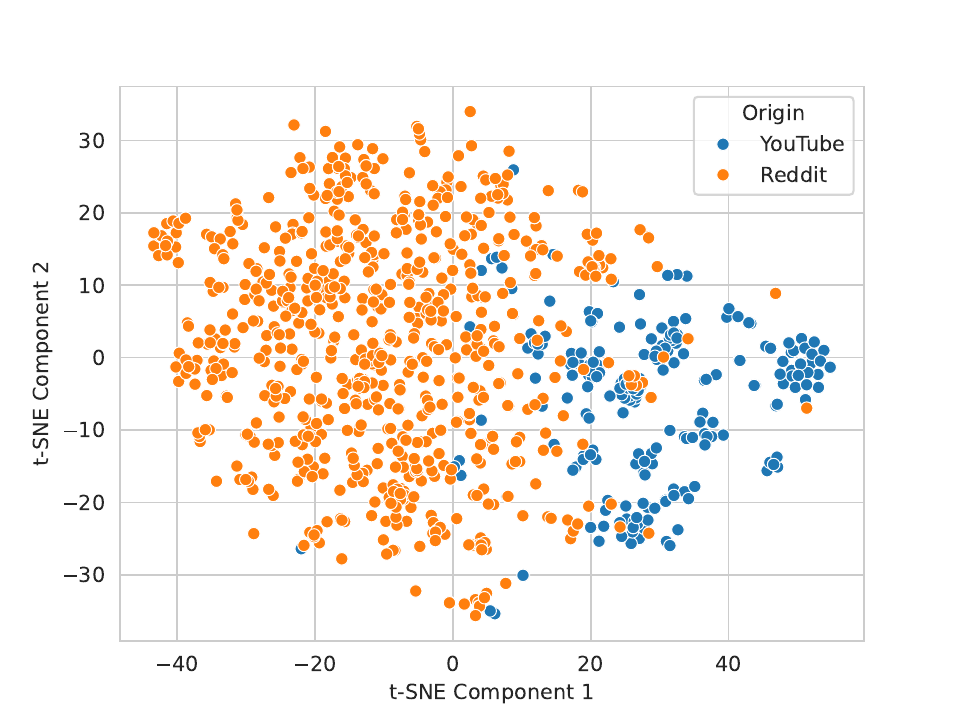}
    \caption{Clozes after dimensionality reduction with $t$-SNE.}
    \label{fig:cloze_text_2d}
\end{figure}

Figure~\ref{fig:hist_outside_tokens_templates} shows the distributions of numbers of outside tokens. 
Figure~\ref{fig:cloze_text_2d} shows our clozes in the two-dimensional plane after conducting $t$-SNE. Table~\ref{tab:annot_examples} shows some of the clozes yielding the highest number of errors per error type.

\definecolor{myblue}{HTML}{1C6DA1}
\definecolor{myorange}{HTML}{E6750C}

\begin{table*}
\centering
\begin{tabular}{@{}lll@{}}
\toprule
%\textbf{Entity} & \textbf{Metric} & \textbf{Cloze Text} \\ \midrule
\multirow{9}{*}{\textbf{Artist}} & \multirow{3}{*}{\textbf{Incorrect}} & \textcolor{myblue}{[Artist] - [WoA] - [Artist] ! ! geroiam [Artist] ! !} \\ 
& & \textcolor{myorange}{rap songs like [Artist]}  \\ 
& & \textcolor{myblue}{[Artist] sings [Artist] ( full album - [Year] )} \\ \cmidrule{2-3}
& \multirow{3}{*}{\textbf{Spurious}} & \textcolor{myblue}{[Artist] - [WoA] ( remix )} \\ 
&  & \textcolor{myorange}{old [Artist] / [Artist] sounding dudes} \\ 
& & \textcolor{myorange}{looking for music that is like [Artist] music} \\ \cmidrule{2-3}
& \multirow{3}{*}{\textbf{Missed}} & \textcolor{myorange}{who is the french [Artist] ?}\\ 
& & \textcolor{myorange}{what is the name of this kind [Artist] ?} \\ 
& & \textcolor{myblue}{[Artist] ( ft . [Artist] ) cover of [WoA] by [Artist]} \\ 
\midrule
\multirow{9}{*}{\textbf{WoA}} & \multirow{3}{*}{\textbf{Incorrect}} & \textcolor{myblue}{[WoA] - [WoA] ( disco version )} \\ 
& & \textcolor{myorange}{hip hop similar to [WoA] from [WoA] album ?} \\ 
& & \textcolor{myorange}{songs / bands like the metalcore [WoA] from [WoA] : waw zombies ?} \\ \cmidrule{2-3}
& \multirow{3}{*}{\textbf{Spurious}} & \textcolor{myblue}{live in central park [ revisited ] : [Artist]} \\ 
& & \textcolor{myblue}{[WoA] - pickin on [Artist] : a bluegrass tribute to [Artist] - pickin on series} \\ 
& & \textcolor{myorange}{similar songs to [Artist] - [WoA] ?} \\ \cmidrule{2-3}
& \multirow{3}{*}{\textbf{Missed}} & \textcolor{myorange}{any songs with some minor [WoA] themes ?} \\ 
& & \textcolor{myblue}{trolls [WoA] comic - con clip | trolls} \\ 
& & \textcolor{myblue}{[Artist] / [Artist] [ [WoA] ] live audio cover} \\ 
\bottomrule
\end{tabular}
\caption{Clozes by data source: \textcolor{myblue}{YouTube} and \textcolor{myorange}{Reddit}.}
\label{tab:outcomes_examples}
\end{table*}

\end{document}